\documentclass[11pt]{article}

\usepackage{arxiv_style}
\usepackage[utf8]{inputenc} %
\usepackage[T1]{fontenc}    %
\usepackage{hyperref}       %
\usepackage{url}            %
\usepackage{booktabs}       %
\usepackage{amsfonts}       %
\usepackage{nicefrac}       %
\usepackage{microtype}      %
\usepackage[dvipsnames]{xcolor} %
\usepackage{amsmath,amssymb,amsthm,mathtools,bm}
\usepackage{graphicx}
\usepackage{subcaption}
\usepackage{algorithm}
\usepackage{algpseudocode}
\usepackage{cleveref}
\usepackage{aliascnt}

\title{Improved Leverage Score Sampling for\\Constrained Active Linear Regression}

\author{\normalsize
  Aarshvi Gajjar\thanks{New York University
    (\href{mailto:aarshvi@nyu.edu}{\nolinkurl{aarshvi@nyu.edu}},
    \href{mailto:adityamakkar@nyu.edu}{\nolinkurl{adityamakkar@nyu.edu}},
    \href{mailto:cmusco@nyu.edu}{\nolinkurl{cmusco@nyu.edu}})}
  \quad
  Syamantak Kumar\thanks{University of Texas at Austin
    (\href{mailto:syamantak@utexas.edu}{\nolinkurl{syamantak@utexas.edu}},
    \href{mailto:joeyzhou@cs.utexas.edu}{\nolinkurl{joeyzhou@cs.utexas.edu}})}
  \quad
  Aditya Makkar\footnotemark[1]
  \quad
  Christopher Musco\footnotemark[1]
  \quad
  Yihan Zhou\footnotemark[2]
}

\newcommand{\RR}{\mathbb{R}}

\newcommand{\PP}{\mathbb{P}}
\newcommand{\EE}{\mathbb{E}}
\renewcommand{\epsilon}{\varepsilon}
\newcommand{\cC}{\mathcal{C}}
\newcommand{\cE}{\mathcal{E}}

\newcommand{\cL}{\mathcal{L}}
\newcommand{\Id}{I}
\newcommand{\opt}{\mathrm{opt}}

\newcommand{\deff}{d_{\mathrm{eff}}}
\newcommand{\dlambda}{d_{\lambda}}
\newcommand{\lev}{\ell}
\newcommand{\tr}{\mathrm{tr}}

\newcommand{\diag}{\mathrm{diag}}
\newcommand{\col}{\mathrm{col}}

\newcommand{\set}[1]{\left\{#1\right\}}
\newcommand{\Set}[2]{\left\{#1:\,#2\right\}}
\newcommand{\paren}[1]{\left(#1\right)}
\newcommand{\bracket}[1]{\left[#1\right]}
\newcommand{\abs}[1]{\left|#1\right|}
\newcommand{\norm}[1]{\left\lVert#1\right\rVert}
\newcommand{\inner}[2]{\left\langle #1, #2 \right\rangle}
\newcommand{\qnorm}[2]{\norm{#1}_{#2}}
\newcommand{\Ellipsoid}[2]{\cE\!\paren{#1,#2}}
\newcommand{\eps}{\varepsilon}
\newcommand{\exampleend}{\hfill\(\lozenge\)}

\DeclareMathOperator*{\argmin}{arg\,min}

\theoremstyle{plain}
\newtheorem{theorem}{Theorem}
\newaliascnt{corollary}{theorem}
\newtheorem{corollary}[corollary]{Corollary}
\aliascntresetthe{corollary}
\newaliascnt{lemma}{theorem}
\newtheorem{lemma}[lemma]{Lemma}
\aliascntresetthe{lemma}
\newaliascnt{proposition}{theorem}

\aliascntresetthe{proposition}
\theoremstyle{definition}
\newaliascnt{definition}{theorem}
\newtheorem{definition}[definition]{Definition}
\aliascntresetthe{definition}
\newaliascnt{remark}{theorem}

\aliascntresetthe{remark}
\newaliascnt{example}{theorem}
\newtheorem{example}[example]{Example}
\aliascntresetthe{example}
\usepackage{thm-restate}

\crefname{theorem}{theorem}{theorems}
\Crefname{theorem}{Theorem}{Theorems}
\crefname{corollary}{corollary}{corollaries}
\Crefname{corollary}{Corollary}{Corollaries}
\crefname{lemma}{lemma}{lemmas}
\Crefname{lemma}{Lemma}{Lemmas}
\crefname{proposition}{proposition}{propositions}
\Crefname{proposition}{Proposition}{Propositions}
\crefname{definition}{definition}{definitions}
\Crefname{definition}{Definition}{Definitions}
\crefname{remark}{remark}{remarks}
\Crefname{remark}{Remark}{Remarks}
\crefname{example}{example}{examples}
\Crefname{example}{Example}{Examples}
\crefname{algorithm}{algorithm}{algorithms}
\Crefname{algorithm}{Algorithm}{Algorithms}

\newcommand{\ellipsoidsdpsection}{
\begingroup

\section{Computing the enclosing ellipsoid}
\label{app:ellipsoid-sdp}

The choice of an enclosing ellipsoid depends on the constraint set and how it is specified.

For finite feasible sets and polytopes specified by vertices, minimizing the total
ellipsoid-regularized leverage-score mass admits a semidefinite formulation.
Let $\cC=\{x_1,\ldots,x_N\}$ or
$\cC=\operatorname{conv}\{x_1,\ldots,x_N\}$, in centered coordinates.
Since an ellipsoid is convex, containment of $\cC$ is equivalent to requiring
$x_i^\top Qx_i\le1$ for every $i\in[N]$.
Thus, writing $B=A^\top A\succ0$ and fixing $\lambda=\lambda_\star\ge0$,
Equation~\eqref{eq:constraint-induced-dimension} becomes
\[
\inf_{Q\succeq0}\ \tr\!\left(B(B+\lambda Q)^{-1}\right)
\quad\text{subject to}\quad
x_i^\top Qx_i\le1,\qquad i\in[N].
\]

This problem has the following semidefinite program formulation:
\begin{equation}
\label{eq:ellipsoid-sdp}
\begin{aligned}
    \inf_{Q,Y}\quad & \tr(Y)\\
    \text{subject to}\quad
        & \begin{pmatrix}
            B+\lambda Q & B^{1/2}\\
            B^{1/2} & Y
          \end{pmatrix}\succeq0,\\
        & x_i^\top Qx_i\le1,\qquad i\in[N],\qquad Q\succeq0.
\end{aligned}
\end{equation}
By the Schur complement, the block constraint is equivalent to
\[
Y\succeq B^{1/2}(B+\lambda Q)^{-1}B^{1/2},
\]
with equality at the optimum for each fixed $Q$.
This applies, for example, to finite feasible sets, simplices, and polytopes
specified by a finite list of vertices.

For more general constraint sets, a minimum-volume enclosing ellipsoid is a possible
heuristic. Computing and approximating such ellipsoids is a well-studied
problem~\citep{CohenCousinsLeeYang2019,WoodruffYasuda2024,GulerGurtuna2012}.

\endgroup
}

\newcommand{\additionalacknowledgments}{%
  \space and Samson Zhou for pointing out the dependence on \(d\) in the lower bound%
}

\AtEndEnvironment{ack}{\par
An earlier version of this paper was developed in 2025 and accepted at the
\href{https://constrained-opt-ml.github.io/}{Constrained Optimization for Machine Learning Workshop}
at NeurIPS that year.
We used GPT iteratively to improve upon some of the proofs developed after the workshop version, as well as to help improve the exposition, obtain feedback on the correctness of the proofs, and assist with the experiments.
}

\date{}

\begin{document}
\maketitle

\begin{abstract}
In this paper we study active learning for constrained linear regression. Given a design matrix $A \in \mathbb{R}^{n \times d}$, a constraint set $\cC \subseteq \mathbb{R}^d$, and query access to a response vector $b \in \mathbb{R}^n$, we seek to output an approximate solution to the constrained least-squares problem $\min_{x\in \cC} \lVert Ax - b\rVert_2^2$ using as few queries to $b$ as possible. This problem arises in many scientific and engineering applications, and \emph{randomized leverage score sampling} provides a powerful method to solve it. However, existing leverage score methods either ignore the constraint or encode it only through ridge regularization, thereby suffering unfavorable query complexity for anisotropic constraint sets. We propose \emph{ellipsoid-regularized leverage scores}, where we first approximate the constraint set by an outer ellipsoid and then use its shape matrix to bias sampling toward directions that matter under the constraint geometry. We prove that the resulting query complexity is governed by a natural \emph{anisotropic effective dimension} $d_{\text{eff}}(\cC)$, which is at most $d$. In particular, with an appropriate choice of parameters, $O(d_{\text{eff}}(\mathcal C)\log d/\varepsilon^2)$ queries to $b$ suffice to return a point in $\cC$ whose objective value is at most $(1+\varepsilon)$ times the optimum. This bound can substantially improve on the standard leverage-score bound, since anisotropic constraints can make $d_{\text{eff}}(\cC)$ much smaller than $d$. Furthermore, for convex constraints, we show how to improve the dependence on $\varepsilon$ from $1/\varepsilon^2$ to $1/\varepsilon$. We complement these upper bounds with two lower bounds: first, a lower bound showing that the $d/\varepsilon^2$ dependence is unavoidable without convexity assumptions on $\cC$, and second, a lower bound showing that, in some regimes, the corresponding effective dimension dependence $\deff(\cC)/\eps$ is necessary for ellipsoidal constraints. We also perform experiments on synthetic regression problems and a parametric heat conduction problem to show that our method achieves lower regression error than standard baselines for similar query budget.

\end{abstract}

\section{Introduction}
\label{sec:introduction}

We study the problem of \emph{active constrained linear regression}. The learner is given a design matrix \(A \in \RR^{n \times d}\) and a feasible set \(\cC \subseteq \RR^d\), but the response vector \(b \in \RR^n\) is initially hidden. The goal is to query as few entries of \(b\) as possible and, using only those observed labels, return a feasible point \(\hat{x} \in \cC\) whose objective value is close to the constrained optimum
\begin{align*}
\opt_\cC \coloneqq \min_{x \in \cC} \cL(x) \qquad \text{where} \qquad \cL(x) \coloneqq \norm{Ax-b}_2^2.
\end{align*}
We consider the agnostic setting, imposing no relation between \(A\) and \(b\), and aim for relative error approximation, i.e., \(\cL(\hat{x}) \le (1+\eps) \opt_\cC\) for some accuracy parameter \(\eps > 0\).

Constrained regression is a classical topic in statistics and optimization, with extensive work on its statistical complexity and computational tractability \citep{AmelunxenLotzMcCoyTropp2014,ChandrasekaranRechtParriloWillsky2012,Chatterjee2014,NegahbanRavikumarWainwrightYu2012,PilanciWainwright2016}.
It arises in applications where constraints encode domain knowledge, including biology \citep{WangParkSusztakZhangLi2019}, imaging and remote sensing \citep{HeinzChang2001,Ozsar_2025}, social science \citep{AbadieDiamondHainmueller2010}, finance \citep{WuYangLiu2014}, and signal processing \citep{LeeBattleRainaNg2006}.

In many applications, unlabeled inputs are plentiful but labels are expensive to acquire, motivating the study of constrained regression in an active learning setting \citep{LookmanBalachandranXueYuan2019,Settles2009,ShieldsGurleyCatarelliChauhanOjedaTuzMasters2023,WarmuthLiaoRaetschMathiesonPuttaLemmen2003,YangZhangChenZhangChen2017}.
One example is surrogate modeling for parametric partial differential equations (PDEs) \citep{AdcockBrugiapagliaWebster2022,CohenDeVore2015}, where active learning is used to approximate maps from PDE parameters to quantities of interest \citep{pmlr-v206-gajjar23a}.
In this setting, each data point \(a_i\) corresponds to a set of PDE parameters, while observing \(b_i\) requires solving the PDE at those parameters and extracting the corresponding quantity of interest.
Thus the design matrix can often be assembled cheaply from candidate parameter values and basis functions, whereas each label may require an expensive numerical simulation.
Natural constraints here include sparsity, coefficient bounds, and positivity of the surrogate's predictions.

\textbf{Existing work on active regression.} Active linear regression is well understood in the unconstrained setting. A canonical non-adaptive approach is leverage-score sampling. The leverage score of the \(i\)-th row \(a_i\) of \(A\) is defined as
\[
\ell_i(A) \coloneqq a_i^\top (A^\top A)^\dagger a_i \ge 0, \qquad i \in [n].
\]
The algorithm samples row indices with probabilities proportional to \(\ell_i(A)\), queries the corresponding entries of \(b\), and solves a reweighted least squares problem on the sampled rows. Since the scores only depend on \(A\), the algorithm can choose all the indices to query in advance, before seeing any entries of \(b\), and then request the labels in parallel. Moreover, leverage score sampling achieves near optimal label complexity \(O(d \log d + d/\eps)\) for a \((1+\eps)\)-relative error approximation of unconstrained regression \citep{Sarlos2006,ChenPrice:2019a}.

Other row-sampling approaches to active regression include volume sampling \citep{DerezinskiWarmuthHsu2018} and related dependent row-sampling schemes \citep{ShimizuChengMusco:2024}.
Beyond linear regression, leverage-score sampling also yields agnostic active-learning guarantees for single-index models with Lipschitz nonlinearities, with label complexity nearly linear in the dimension for fixed accuracy \citep{pmlr-v247-gajjar24a}.

\textbf{Taking constraints into account.} The same leverage score sampling also gives a baseline for constrained regression. A result in \citet{ClarksonWoodruff2017} implies that \(O(d\log(d)/\eps^2)\) queries suffice to obtain a \((1+\eps)\)-relative error solution to \( \min_{x\in \cC}\|Ax-b\|_2^2\) for an arbitrary feasible set \(\cC\); we make this implication precise in \Cref{sec:preliminaries}. However, this bound does not depend on the geometry of \(\cC\): it is unchanged for a Euclidean ball, an elongated ellipsoid, or a more restricted feasible set.

This contrasts with random-projection methods for convex constraints, where the constraint is known to affect the required projection dimension. \citet{PilanciWainwright2015} show that the projection dimension is governed by the tangent cone of \(\cC\) at the constrained optimizer. Related iterative sketching methods for constrained least squares repeatedly use sketched objectives, either through lower-dimensional subproblems or projected updates \citep{PilanciWainwright2016,TangGolbabaeeDavies2017}. These results point to the fact that, under random projections, constraints can control the effective dimension of the problem. These projection-dimension guarantees do not directly yield label-query bounds, since forming a dense projection of the response vector generally requires access to all its entries. \emph{We ask whether an analogous, constraint-dependent improvement is possible in the active row-sampling model.}

\textbf{Ridge leverage scores.} A natural baseline for incorporating constraints into row sampling is ridge leverage-score sampling \citep{AlaouiMahoney2015}, which has been used in kernel ridge regression, regularized data fitting, and low-rank approximation \citep{AvronClarksonWoodruff2017,CohenMuscoMusco2017}.
For \(\lambda>0\), the \(\lambda\)-ridge leverage score of row \(a_i\) is
\[
\ell_i^\lambda(A) \coloneqq a_i^\top (A^\top A + \lambda \Id)^{-1} a_i \ge 0.
\]
These scores incorporate regularization via an isotropic penalty, i.e., the matrix \(\lambda I\) penalizes all directions in the parameter space equally. Equivalently, ridge
regularization, imposed through the objective \(
    \min_x \|Ax-b\|_2^2+\lambda\|x\|_2^2
\)
is the Lagrangian relaxation of least squares constrained to a Euclidean ball for some choice of \(\lambda\). Thus when used as a proxy for a general feasible set, ridge leverage scores encode information about \(\cC\) by a spherical approximation.

The spherical approximation is well matched when \(\cC\) is roughly isotropic, but it can be overly conservative for anisotropic constraints, i.e., when \(\cC\) has different scales in different directions. \phantomsection\label{ex:intro-motivating} For instance, consider \(A = I_2\) and \(\cC = \{x \in \RR^2 : x_1^2 + 100 x_2^2 \le 1\}\). Then ridge leverage scores assign equal weight to the two rows, since \(\ell_1^\lambda(A) = \ell_2^\lambda(A) = 1/(1+\lambda)\). The constraint, however, treats the two coordinates very differently: it allows \(x_1 \in [-1,1]\) while restricting \(x_2 \in [-0.1,0.1]\). For illustration, suppose we seek an additive excess loss of at most \(10^{-2}\) and know in advance that \(b_2 \in [-0.1,0.1]\). Then the second label need not be queried at all. Indeed, for any feasible \(x=(x_1,x_2)\), the point \((x_1,0)\) is also feasible, and replacing \(x_2\) by \(0\) increases the loss by at most \(b_2^2\le 10^{-2}\). Therefore, in this simple example, the second coordinate is already controlled by the constraint, and it may be important to leverage this fact to save label queries.

\textbf{Our proposed scores.} The preceding example motivates a constraint-dependent analogue of ridge leverage scores. Instead of regularizing all directions equally through the penalty \(\lambda \Id\), we introduce a positive definite matrix that reflects the directional structure of the feasible set. Let \[
\Ellipsoid{c}{Q}
\coloneqq
\Set{x \in \RR^d}{(x-c)^\top Q (x-c) \le 1},
\qquad Q \succ 0,
\]
be an outer ellipsoid for \(\cC\), centered at $c \in \RR^d$ so that \(\cC \subseteq \Ellipsoid{c}{Q}\). We define the corresponding \emph{ellipsoid-regularized leverage score} by
\[
\ell_i^{\lambda,Q}(A) \coloneqq a_i^\top (A^\top A + \lambda Q)^{-1} a_i \ge 0.
\]

Here, the directions that are more tightly restricted by the constraint are penalized more strongly. Similar to leverage score sampling, we sample rows from \(A\) with probabilities proportional to these scores. The total sampling mass for this sampling rule is
\[
\dlambda(A;Q)
\coloneqq
\sum_{i=1}^n a_i^\top (A^\top A + \lambda Q)^{-1} a_i,
\]
a quantity which is the analogue of the score mass used in ridge leverage score sampling. This total mass governs the sample complexity for the corresponding sampling algorithm. Optimizing over admissible outer ellipsoids yields the constraint-induced effective dimension \(\deff(\cC)\) that appears in our main bounds. At a fixed regularization level, replacing an outer ellipsoid by a contained ellipsoid with the same center can only reduce the total score mass. We illustrate this effect in \Cref{ex:anisotropic-identity}.

Our main results build on this observation: ellipsoid-regularized sampling followed by weighted constrained least squares gives an excess-loss guarantee with query complexity controlled by \(\deff(\cC)\), up to logarithmic and accuracy factors. To the best of our knowledge, this is the sharpest known dimension dependence for non-adaptive constrained active regression, improving over ambient-dimension bounds.

\subsection{Our contributions}
The approximation guarantees below hold with constant success probability.

\textbf{Regression guarantees for arbitrary constraints.} We analyze ellipsoid-regularized leverage score sampling for arbitrary feasible constraint sets. \Cref{thm:general-constraint-aware} shows that sampling
\(
m =
\tilde{O}\!\paren{
    \deff(\cC)/\eps^2
}
\)
labels suffices to obtain a \((1+\eps)\)-approximate constrained regression solution.\footnote{\(\tilde{O}\) hides logarithmic factors in \(d\).}

\textbf{Improved accuracy dependence for convex constraints.} When \(\cC\) is closed and convex, we give a tighter analysis of our ellipsoid-regularized sampling algorithm in \Cref{thm:convex-ellipsoid-regularized}, showing a sample complexity of \(\tilde{O}(\deff(\cC)/\eps)\) for the same relative-error guarantee. This improves the accuracy dependence from \(1/\eps^2\) to \(1/\eps\) while retaining dimension dependence governed by \(\deff(\cC)\). Notably, in \Cref{thm:convex-projection}, we also show that a simple algorithm based on standard leverage score sampling and projection obtains sample complexity \(\tilde{O}(d/\eps)\), matching the accuracy dependence of unconstrained regression.

\textbf{Lower bounds.} We prove two complementary lower bounds demonstrating that the main dependencies in our upper bounds are necessary in the worst case. First, for non-convex constraints, we adapt the biased-coin construction of \citet[Theorem~13]{ChenDerezinski2021} to show in \Cref{thm:lb-d-eps-assouad} that any algorithm, even adaptive, requires \(\Omega(d/\eps^2)\) label queries, matching the \(1/\eps^2\) dependence of our general guarantee in the worst case. Second, building on the framework of \citet{ChenPrice:2019a} and \citet[Theorem~4]{DevvritRajaramanAwasthi2021v1}, we show in \Cref{cor:effective-dimension-ball} that the appropriate ellipsoid-regularized sampling mass must appear with a \(1/\eps\) factor for ellipsoidal constraints in a natural target-accuracy regime.

\textbf{Experiments.} In \Cref{sec:experiments}, we evaluate the sample efficiency of our method on synthetic data and a parametric heat equation. We compare our method with uniform, leverage-score, and ridge leverage-score sampling using the same number of samples.

\section{Preliminaries}
\label{sec:preliminaries}

\textbf{Notations.} We write \([n] \coloneqq \set{1,\dots,n}\). The \(i\)-th row of \(A\) is denoted by \(a_i\), where \(a_i \in \RR^d\) is viewed as a column vector. We write \(e_i\) for the \(i\)-th standard basis vector, whose only nonzero entry is a \(1\) in position \(i\). Throughout the paper, we assume that \(A\) has full column rank, so \(A^\top A\) is invertible. For a matrix \(M\), \(\tr(M)\) denotes its trace. We also write
\(
\Id_n
\)
for the \(n \times n\) identity matrix. For symmetric matrices \(M,N\), we write \(M\succeq N\) in the Löwner order when \(M-N\) is positive semidefinite. For a positive-definite matrix \(Q \in \RR^{d\times d}\), we write
\(
\qnorm{u}{Q} \coloneqq \sqrt{u^\top Q u}.
\)
We write the least-squares loss as
\(
\cL(x) \coloneqq \norm{Ax-b}_2^2.
\)
We assume throughout that \(\cC\) is nonempty and closed. Since \(A\) has full
column rank, \(\cL(x)=\|Ax-b\|_2^2\) is coercive, and hence
\(\min_{x\in\cC}\cL(x)\) is attained.

\textbf{Row sampling model.} A typical randomized row-sampling procedure is represented by a sampling-and-rescaling matrix.
Let \(p=(p_1,\ldots,p_n)\) be a probability vector on \([n]\), and let \(m\) be the number of samples.
We draw indices \(J_1,\ldots,J_m\) independently with replacement from \([n]\), where \(\Pr[J_t=i]=p_i\).
We define \(S\in\RR^{m\times n}\) row by row by
\begin{equation}
\label{eq:sampling-matrix}
S_{t,\cdot} = \frac{1}{\sqrt{m p_{J_t}}}\,e_{J_t}^{\top}.
\end{equation}
Therefore \(SA\) consists of the sampled and rescaled rows \((m p_{J_t})^{-1/2}a_{J_t}^{\top}\), and \(Sb\) consists of the corresponding queried and rescaled labels \((m p_{J_t})^{-1/2}b_{J_t}\).
The factor \(p_{J_t}^{-1/2}\) corrects for unequal sampling probabilities: since row \(i\) appears \(m p_i\) times in expectation, each appearance is given squared weight \(1/(m p_i)\), so the expected total contribution of row \(i\), for \(p_i>0\), equals its contribution in the original loss.

For active constrained least squares, a starting point is leverage score sampling, where we choose these probabilities \(p_i\) proportional to the leverage scores of the rows of \(A\) which are defined as follows.

\begin{definition}[Leverage score]
\label{def:plain-leverage}
Let \(U \in \RR^{n \times d}\) be a matrix whose columns form an orthonormal basis for the column space of \(A \in \RR^{n \times d}\). Let \(a_i\) and \(u_i\) be the \(i\)-th rows of \(A\) and \(U\), respectively. The leverage score \(\ell_i\) of row \(i\) of \(A\) is
\[
\ell_i(A)
\coloneqq
\norm{u_i}_2^2
=
a_i^\top (A^\top A)^{-1} a_i
=
\max_{x \in \RR^d:\,x\ne0}
\frac{(a_i^\top x)^2}{\norm{Ax}_2^2}.
\]

\end{definition}

Since \(A\) has full column rank, \(0 \le \ell_i(A) \le 1\), and
\(
\sum_{i=1}^n \ell_i(A) = d.
\) These scores depend only on \(A\), so the sampling distribution can be chosen before any responses are observed.

\textbf{Revisiting leverage score sampling for constrained regression.} Using leverage score sampling, one can already obtain a good approximation to the constrained regression problem using \(O(d\log d/\eps^2)\) samples. The key ingredient is an affine embedding, which states that the sampled objective, \(\norm{S(Ax-b)}_2^2\), approximates the original objective, \(\cL(x)\), uniformly over \(x \in \RR^d\), up to an additive error. This affine embedding guarantee follows from \citep[Theorems~39 and~42]{ClarksonWoodruff2017}, and we restate it in \Cref{lem:plain-affine-embedding} below.

\begin{lemma}[Affine embedding]
\label{lem:plain-affine-embedding}
Let \(A \in \RR^{n \times d}\), and let \(b \in \RR^n\). Let \(S\) be the \(m \times n\) sampling-and-rescaling matrix defined by \eqref{eq:sampling-matrix} with sampling probabilities \(p_i = \ell_i(A)/d\).

There is a constant \(C > 0\) such that, for every \(\eps \in (0,1/2)\), if \(m \ge C\,\frac{d\log(2d)}{\eps^2}\), then with constant probability, there exists a scalar \(\Delta=\Delta(S,b)\) such that for every \(x \in \RR^d\),
\begin{equation}
\label{eq:plain-affine-embedding}
(1-\eps)\cL(x)
\le
\norm{S(Ax-b)}_2^2-\Delta
\le
(1+\eps)\cL(x).
\end{equation}
\end{lemma}
One  can choose \(\Delta = \norm{Sz}_2^2 - \norm{z}_2^2\), where the vector \( z \coloneqq \paren{\Id_n - A(A^\top A)^{-1}A^\top}b.\) Since $\Delta$ is independent of $x$, subtracting it does not change the minimizer of the sampled objective over any feasible set. Moreover, since \eqref{eq:plain-affine-embedding} holds uniformly for all $x \in \RR^d$, it directly gives the following regression guarantee for leverage score sampling. 

\begin{corollary}
\label{cor:baseline-constrained}
Under the assumptions of \Cref{lem:plain-affine-embedding}, if \(\hat{x}_{\mathrm{lev}} \in \argmin_{x\in\cC}\norm{S(Ax-b)}_2^2\), then, with constant probability,
\[
\cL(\hat{x}_{\mathrm{lev}}) \le (1+4\eps)\min_{x \in \cC} \cL(x) .
\]
\end{corollary}
\emph{Proof.}
Let \(x_\cC^\star\in\argmin_{x\in\cC}\cL(x)\). By \Cref{lem:plain-affine-embedding} and the optimality of \(\hat{x}_{\mathrm{lev}}\), we obtain:
\[
(1-\eps)\cL(\hat{x}_{\mathrm{lev}}) \le \norm{S(A\hat{x}_{\mathrm{lev}}-b)}_2^2-\Delta \le \norm{S(Ax_\cC^\star-b)}_2^2-\Delta \le (1+\eps)\cL(x_\cC^\star).
\]Rearranging the above gives the desired guarantee. \qed

\section{Constraint-informed sampling}
\label{sec:constraint_aware_sampling}

\Cref{cor:baseline-constrained} shows that standard leverage-score sampling provides a guarantee for arbitrary feasible sets, but its sampling distribution depends only on \(A\) and does not use the geometry of \(\cC\). In a constrained problem, this might be too conservative: the feasible set \(\cC\) may restrict some directions more heavily than others, and therefore a good sampling method should take that into account. The following example illustrates this limitation.

\begin{example}[A constrained limitation of leverage scores]\label{ex:anisotropic-identity}
Consider \(A = [\Id_d;0] \in \RR^{n\times d}\), so each non-zero row of $A$ observes only one coordinate. Consequently, leverage scores assign equal mass to every nonzero row, i.e., \(\ell_i(A)=1\) for \(i \in [d]\).

Now let $\cC$ be the axis-aligned ellipsoid \(\cC=\Set{x\in\RR^d}{\sum_{k=1}^d x_k^2/R_k^2\le 1}\), where \(Q=\diag(R_1^{-2},\dots,R_d^{-2})\) and where the radii \(R_1,\dots,R_d > 0\) may be highly nonuniform. Since row $i$ of $A$ only observes coordinate $i$, its feasible prediction range is $\{a_i^\top x : x\in\cC\} = [-R_i,R_i]$. Thus, some coordinates can be much more constrained than others, and standard leverage scores do not capture this distinction by assigning the same score to each nonzero row.
\exampleend

\end{example}

\subsection{Ellipsoid-regularized leverage scores}

The example suggests a sampling method that can distinguish between directions that are highly and weakly constrained by the feasible set. To see how to incorporate this information, we recall the maximization characterization of leverage scores in \Cref{def:plain-leverage}:
\(
\ell_i(A)
=
\max_{x \in \RR^d:\,x\ne0}
\frac{(a_i^\top x)^2}{\norm{Ax}_2^2}.
\)
This quantity measures how large the prediction \(a_i^\top x\) can be relative to the overall prediction \(\norm{Ax}_2\) across all rows. If \(\cC \subseteq \Ellipsoid{c}{Q}\), we first recenter the regression problem at \(c\).\footnote{The center \(c\) can be removed by the change of variables \(x=c+u\): setting \(\cC_c=\cC-c\) and \(b_c=b-Ac\), the problem \(\min_{x\in\cC}\cL(x)\) is equivalent to \(\min_{u\in\cC_c}\norm{Au-b_c}_2^2\), with \(\cC_c\subseteq\cE(Q)\).}
For the rest of the paper, we work in centered coordinates, relabeling the variable, feasible set, and response vector as \(x\), \(\cC\), and \(b\), respectively, so that \(\cC\subseteq\cE(Q)\).
The quantity \(x^\top Qx\) then quantifies how strongly the ellipsoid restricts movement away from the origin. To incorporate this geometry, we replace the denominator \(\norm{Ax}_2^2\) with \(\norm{Ax}_2^2+\lambda x^\top Qx\), where \(\lambda\ge0\) controls how strongly the outer ellipsoid affects the scores.

This leads to the definition of ellipsoid-regularized leverage scores. 
\begin{definition}[Ellipsoid-regularized leverage score]
\label{def:ellipsoid-regularized-leverage}
Let \(A \in \RR^{n\times d}\), let \(Q \succ 0\), and let \(\lambda\ge0\).
For row \(i\), the \((\lambda,Q)\)-ellipsoid-regularized leverage score is
\begin{equation}
\label{eq:ellipsoid-lev-score}
\lev_i(A;\lambda,Q)
\coloneqq
\max_{x \in \RR^d:\,x\ne0}
\frac{(a_i^\top x)^2}{\norm{Ax}_2^2+\lambda x^\top Qx}
=
a_i^\top (A^\top A + \lambda Q)^{-1} a_i.
\end{equation}
\end{definition}

Using these scores, the sample complexity is governed by the total score mass \(\dlambda(A;Q)\):
\begin{equation}
\label{eq:regularized-score-mass}
\dlambda(A;Q)
\coloneqq
\sum_{i=1}^n \lev_i(A;\lambda,Q)
=
\tr\!\paren{A(A^\top A + \lambda Q)^{-1} A^\top}.
\end{equation}

Since \(A^\top A+\lambda Q\succeq A^\top A\), we have \(0\le\lev_i(A;\lambda,Q)\le\ell_i(A)\) for every \(i\), and hence \(\dlambda(A;Q)\le\sum_{i=1}^n\ell_i(A)=d\).

Note that when \(\lambda=0\), these are the standard leverage scores. The impact of $Q$ becomes visible once $\lambda >0$. Returning to \Cref{ex:anisotropic-identity}. Suppose that \(A^\top A=\Id_d\) and \(Q=\diag(R_1^{-2},\ldots,R_d^{-2})\), then for \(i\in[d]\) the ellipsoid-regularized leverage score is:
\[
\lev_i(A;\lambda,Q)
=
e_i^\top(\Id_d+\lambda Q)^{-1}e_i
=
(1+\lambda/R_i^2)^{-1}.
\]
Hence \(
\dlambda(A;Q) = \sum_{i=1}^d (1+\lambda/R_i^2)^{-1}.\) When only \(k\) squared radii are large relative to \(\lambda\) and the remaining scores sum to much less than \(k\), then \(\dlambda(A;Q) \approx k\) even though the ambient dimension is \(d\), capturing the gap not identified by leverage scores.

\noindent\textbf{Connection to ridge leverage scores.} When the outer ellipsoid is the Euclidean ball of radius \(R\), its shape matrix is \(Q=R^{-2}\Id_d\). In this case, \(\lev_i(A;\lambda,Q) = a_i^\top (A^\top A + \lambda R^{-2}\Id_d)^{-1} a_i\), recovering the ridge leverage scores proposed in \citep{AlaouiMahoney2015} with ridge parameter \(\lambda/R^2\).  

\noindent\textbf{On choosing $Q$}. The choice of $Q$ remains important to govern the sample complexity. A simple monotonicity argument gives a partial ordering among admissible choices. If \(Q_1 \succeq Q_2\), then \(A^\top A + \lambda Q_1 \succeq A^\top A + \lambda Q_2\), implying that \((A^\top A + \lambda Q_1)^{-1} \preceq (A^\top A + \lambda Q_2)^{-1}\), and hence \(\lev_i(A;\lambda,Q_1) \le \lev_i(A;\lambda,Q_2)\) for every row \(i\), while the total score satisfies \(\dlambda(A;Q_1)\le \dlambda(A;Q_2)\).

Therefore, among admissible outer ellipsoids with comparable shape matrices, the better choice for sampling is the one whose shape matrix is larger in the Löwner order. This does not determine the optimal \(Q\), since shape matrices may be incomparable in Löwner order.

\subsection{Constrained regression from ellipsoid-regularized sampling}

The sampling procedure is described in \Cref{alg:ellipsoid-regularized-sampling}: for a fixed matrix $Q$ and a chosen \(\lambda \ge0\), we sample rows of $A$ with probabilities proportional to \(\ell_i(A;\lambda,Q)\) and solve the constrained least squares problem on the sampled instance. 

\begin{algorithm}[t]
\caption{Ellipsoid-regularized leverage-score sampling}
\label{alg:ellipsoid-regularized-sampling}
\begin{algorithmic}[1]
\Statex \textbf{Given:} Matrix \(A\), query access to \(b\), feasible set \(\cC\subseteq\cE(Q)\), \(\lambda\ge0\), sample size \(m\).
\State Set \(p_i=\lev_i(A;\lambda,Q)/\dlambda(A;Q)\) for \(i\in[n]\).
\State Draw \(m\) i.i.d.\ indices from \(p\), form the sampling-and-rescaling matrix \(S\) as in \eqref{eq:sampling-matrix}, and query the corresponding entries of \(b\).
\State Compute
\(
\hat{x}\in\argmin_{x\in\cC}\norm{S(Ax-b)}_2^2.
\)
\State \Return \(\hat{x}\).
\end{algorithmic}
\end{algorithm}

For a relative-error guarantee, an appropriate parameter value will be
\(
\lambda_\star \le c_\lambda\frac{\eps}{1+\eps}\opt_\cC,
\)
for a sufficiently small \(c_\lambda>0\).\footnote{Throughout, we assume knowledge of a constant-factor approximation to \(\opt_\cC\), which is used to select \(\lambda_\star\). If such an estimate is not available, the additive form of the guarantee still holds: for any chosen \(\lambda\ge0\), we obtain \(\cL(\hat{x})\le(1+\eps)\opt_\cC+O(\lambda)\), after adjusting the absolute constant in the sample size.}
For bounded \(\cC\)\footnote{Unbounded sets can be accommodated by allowing \(Q\succeq0\) with \(\cC\subseteq\cE(Q)\).}, we define a constraint-dependent effective dimension at this choice of \(\lambda_\star\) by optimizing the shape matrix \(Q\) while holding the center fixed. In the centered coordinates introduced above, this gives
\begin{equation}
\label{eq:constraint-induced-dimension}
\deff(\cC)
\coloneqq
\inf_{Q\succ0}
d_{\lambda_\star}(A;Q)
\qquad
\text{subject to}
\qquad
\cC \subseteq \cE(Q).
\end{equation}
The notation \(\deff(\cC)\) suppresses the dependence on \(A\) and \(\lambda_\star\). The response vector \(b\) and target accuracy \(\eps\) enter through the choice of \(\lambda_\star\). We hold these quantities and the ellipsoid center fixed when optimizing over \(Q\).

The infimum in \eqref{eq:constraint-induced-dimension} need not be attained, but a constant-factor approximation suffices for our guarantees. When \(\cC\ne\{0\}\), we have \(\deff(\cC)>0\), so we can choose a feasible \(Q\succ0\) satisfying \(d_{\lambda_\star}(A;Q)\le2\deff(\cC)\). We use such a \(Q\) in the following guarantees and absorb the factor of two into the constants in the query bounds. If \(\cC=\{0\}\), we simply return \(0\) without querying \(b\).

When \(\cC\) is given explicitly as a finite point set or its convex hull, Appendix~\ref{app:ellipsoid-sdp} gives a semidefinite program whose infimum equals \(\deff(\cC)\).

We now state the regression guarantee with sample complexity determined by this effective dimension.

\begin{restatable}{theorem}{generalconstraintawarethm}
\label{thm:general-constraint-aware}
Let \(A\in\RR^{n\times d}\), \(b\in\RR^n\), \(\cC\subseteq\RR^d\), \(\eps \in (0,1/2)\), and \(\delta \in (0,1)\). Let \(Q\succ0\) be chosen as described in \eqref{eq:constraint-induced-dimension}, and let $\hat{x}$ be the output of \Cref{alg:ellipsoid-regularized-sampling} with \(Q\) and \(\lambda_\star\). There is a constant $C>0$ such that, if
\(
m
\ge
C\left[
\frac{\deff(\cC)}{\eps^2}\log\left(\frac{d}{\delta}\right)
+
\frac{\deff(\cC)}{\eps^2 \delta}
\right],
\)
then, with probability at least
\(1-\delta\),
\[
\cL(\hat{x})
\le
(1+\eps)\opt_\cC .
\]
\end{restatable}
\textbf{Proof sketch.}
The proof follows the standard active regression analysis for ridge leverage score sampling. There are two probabilistic ingredients: the first is a regularized analogue of a subspace-embedding guarantee, as shown in \Citep{AlaouiMahoney2015}, according to which, setting \(K = A^\top A + \lambda_\star Q\), we obtain with high probability for every \(z \in \RR^d\),
\[\norm{A z}_2^2 - \norm{SAz}_2^2 \le \eps\norm{A z}_2^2 + \lambda_\star z^\top Q z.\]
The second ingredient is an approximate matrix multiplication bound for a fixed residual \(r = A x_0 - b\) where \(x_0 \in \argmin_{x\in\cC}\cL(x)\), which controls with high probability \(K^{-1/2} A^\top (S^\top S - I)r\) using a second moment argument. Since \(\hat{x}\) and \(x_0\) both lie in \(\cC\subseteq\cE(Q)\), their difference satisfies \(\qnorm{\hat{x}-x_0}{Q}\le2\). This bounds the regularization term introduced by the embedding guarantee. Conditioning on these two events, the optimality of \(\hat{x}\) for the sampled objective gives the desired guarantee. Full details can be obtained from \Cref{app:constraint-aware-proofs}.

\subsection{Improved bounds for convex constraints}

The sample complexity in the preceding result has a \(1/\eps^2\) dependence. This dependence is unavoidable for arbitrary constraints; we establish this in \Cref{thm:lb-d-eps-assouad} via a lower bound. Interestingly, for convex constraints, we show that using an algorithm based on leverage score sampling, one can recover a sharper \(1/\eps\) dependence. The algorithm, described in \Cref{alg:convex-projection}, separates the sampling step from the feasibility step. The simple idea is to first solve the unconstrained \emph{sampled} least-squares objective and then project \(A\hat{x}_{\mathrm u}\) onto \(A\cC=\{Ax:x\in\cC\}\). Under the standing assumptions, if \(\cC\) is convex, then \(A\cC\) is closed and convex, so the projection onto \(A\cC\) is well-defined.

\begin{algorithm}[t]
\caption{Leverage-score sampling for convex constraints}
\label{alg:convex-projection}
\begin{algorithmic}[1]
\Statex \textbf{Given:} Matrix \(A\), query access to \(b\), feasible set \(\cC\), sample size \(m\).
\State Set \(p_i=\ell_i(A)/d\) for \(i\in[n]\).
\State Draw \(m\) i.i.d.\ indices from \(p\), form the sampling-and-rescaling matrix \(S\) as in \eqref{eq:sampling-matrix}, and query the corresponding entries of \(b\).
\State Compute
\(
\hat{x}_{\mathrm{u}}\in\argmin_{x\in\RR^d}\norm{S(Ax-b)}_2^2.
\)
\State Compute
\(
\hat{x}_{\mathrm{proj}}\in\argmin_{x\in\cC}\norm{A\hat{x}_{\mathrm{u}}-Ax}_2^2.
\)
\State \Return \(\hat{x}_{\mathrm{proj}}\).
\end{algorithmic}
\end{algorithm}

\begin{restatable}[Improved bound for convex constraints]{theorem}{convexprojectionthm}
\label{thm:convex-projection}
Let \(\cC\subseteq\RR^d\) be convex. Fix \(\epsilon\in(0,1)\), set
\(
p_i=\frac{\ell_i(A)}{d},
\) and draw \(m\) rows independently from \(p\) to form the sampling-and-rescaling matrix \(S\). Let
\[
\hat{x}_{\mathrm{u}} \in \argmin_{x\in\RR^d}\norm{S(Ax-b)}_2^2,
\qquad
\hat{x}_{\mathrm{proj}}
\in
\argmin_{x\in\cC}\norm{A\hat{x}_{\mathrm{u}}-Ax}_2^2.
\]
There is an absolute constant \(C>0\) such that, if
\(m
\ge
C\left(
d\log d
+
\frac{d}{\epsilon}
\right),
\)
then with constant probability,
\(
\cL(\hat{x}_{\mathrm{proj}})
\le
(1+\epsilon)\opt_\cC.
\)
\end{restatable}

We can in fact obtain a stronger guarantee for convex constraints using ellipsoid-regularized sampling. Namely, when \(\cC\) is convex, \Cref{alg:ellipsoid-regularized-sampling} achieves both the sharper \(1/\eps\) dependence and a dimension dependence governed by \(\deff(\cC)\), without any projection step.

\begin{restatable}[Improved ellipsoid-regularized bound for convex constraints]{theorem}{convexellipsoidregularizedthm}
\label{thm:convex-ellipsoid-regularized}
Let \(A\in\RR^{n\times d}\), \(b\in\RR^n\), \(\cC\subseteq\RR^d\), \(\eps \in (0,1/2)\), and \(\delta \in (0,1)\). Assume that \(\cC\) is convex. Let \(Q\succ0\) be chosen as described in \eqref{eq:constraint-induced-dimension}, and let \(\hat{x}\) be the output of \Cref{alg:ellipsoid-regularized-sampling} with \(Q\) and \(\lambda_\star\). There is a constant \(C>0\) such that, if
\(
m
\ge
C\left[
\deff(\cC)\log\left(\frac{d}{\delta}\right)
+
\frac{\deff(\cC)}{\eps\delta}
\right],
\)
then, with probability at least \(1-\delta\),
\(
\cL(\hat{x})
\le
(1+\eps)\opt_\cC .
\)
\end{restatable}

The proof of \Cref{thm:convex-ellipsoid-regularized} uses a sharper analysis of \Cref{alg:ellipsoid-regularized-sampling} for convex sets. For convex constraints, the squared prediction error relative to the constrained optimum is at most the excess loss. This allows the sampled gradient to have an error of order \(\sqrt{\eps}\), rather than \(\eps\), leading to the \(1/\eps\) sample dependence. We defer the details to \Cref{app:convex-projection-proofs}.

\section{Lower bounds}
\label{sec:lower_bounds}
We next give lower bounds for two types of constraint sets:
\begin{enumerate}
\item For general nonconvex constraints,
\(\Omega(d/\eps^2)\) queries are unavoidable in the worst case.
\item For ellipsoidal constraints and fixed \((\lambda,Q)\),
obtaining a \((1+\eps)\)-approximate solution requires
\(\Omega(\dlambda(A;Q)/\eps)\) label queries in the worst case,
for sufficiently small \(\eps\).
\end{enumerate}
Both proofs use Yao's minimax principle together
with information-theoretic
arguments and a version of the weighted Assouad lemma to convert transcript
indistinguishability into sign-recovery error. The difference between the two
constructions is how sign errors translate into excess loss. Detailed proofs are deferred to
\Cref{app:lower-bound-proofs}.

\subsection{Lower bound for nonconvex feasible sets}
\label{subsec:lower-bound-nonconvex}
We adapt the construction of \citet[Theorem~13]{ChenDerezinski2021}
for unconstrained least absolute deviation regression. Here, we use squared loss and
a nonconvex feasible set.

The hard instance uses the hypercube \(\set{-1,1}^d\) and a repeated-identity
design. Each coordinate encodes an unknown sign in the bias of a coin, and achieving
relative error requires recovering sufficiently many of these signs.
Unconstrained active least squares admits an \(O(d/\eps)\) query bound
\citep{ChenPrice:2019a}, whereas general nonconvex constraints can require
\(\Omega(d/\eps^2)\) queries. Thus, the \(1/\eps\) accuracy dependence available for
convex constraints in \Cref{thm:convex-projection} cannot extend to arbitrary feasible sets.

\begin{restatable}{theorem}{nonconvexlowerboundthm}
\label{thm:lb-d-eps-assouad}
There exist absolute constants \(c,C>0\) such that the following holds. Fix
\(\eps \in (0,1/400]\) and \(d \ge 2\), and set
\(n \coloneqq \left\lceil \frac{C\log(100d)}{\eps^2}\right\rceil.\) Then there exist a matrix \(A \in \RR^{nd\times d}\) and a nonconvex feasible set \(
\cC = \set{-1,1}^d
\) for which any randomized active algorithm that queries fewer than
\(c\,d/\eps^2\) entries of \(b\) and returns \(\hat{x} \in \cC\) must fail on some
response vector \(b \in \RR^{nd}\) in the sense that
\[
\PP_{\mathrm{alg}}\!\left[
\cL(\hat{x})
\le
(1+2\eps)\opt_\cC
\right]
< 0.95.
\]
\end{restatable}

\subsection{Lower bound for ellipsoidal constraints}

For ellipsoidal constraints, our lower bound involves a truncated effective dimension,
retaining only eigenvalues at least \(\eps\).

\begin{definition}[Truncated effective dimension]
\label{def:truncated-effective-dimension}
Let \(H=A^\top A+\lambda Q\succ0\), and let \(\rho_1,\ldots,\rho_d\) be the eigenvalues
of \(H^{-1/2}A^\top A H^{-1/2}\). For \(\alpha>0\), define
\(\dlambda^\alpha(A;Q)\coloneqq\sum_{j=1}^d
\rho_j\,\mathbf{1}\set{\rho_j\ge\alpha}\).
Thus \(\dlambda(A;Q)=\sum_j\rho_j\) is the effective dimension for fixed \(Q\), while
\(\dlambda^\alpha\) keeps only directions
whose regularized leverage eigenvalue is at least the threshold \(\alpha\).
\end{definition}

For eigenvalues below \(\eps\), regularization can shrink the solution so much that
sign errors have little effect on the loss. Our construction therefore
uses only directions with eigenvalues at least \(\eps\).

The framework was originally developed by \citet{ChenPrice:2019a}, and we use the
formulation for active ridge regression from \citet[Theorem~4]{DevvritRajaramanAwasthi2021v1}.
In particular, we strengthen the concentration argument by specifying a finite number
of row repetitions. We also extend the proof from an isotropic to a diagonal quadratic
regularizer and reprove the sign-recovery step.

We first establish a lower bound for quadratically regularized regression, then
transfer it to regression over a fixed ellipsoid.

\begin{restatable}{theorem}{effectivedimensionlowerboundthm}
\label{thm:effective-dimension-lower-bound}
There exist absolute constants \(c,c_0,C>0\) such that the following holds. Fix
\(d\ge2\), \(\lambda\ge0\), a diagonal matrix \(Q=\diag(q_1,\ldots,q_d)\) with
\(q_i>0\), and \(\eps\in(0,1]\). Set
\(n=\lceil C\log(100d)/\eps\rceil\). Then there exists a matrix
\(A\in\RR^{nd\times d}\) with \(A^\top A=\Id_d\) such that, for
\[
\cL_{\lambda,Q}(x;b)\coloneqq\norm{Ax-b}_2^2+\lambda x^\top Qx,
\qquad
\opt_{\lambda,Q}(A,b)\coloneqq\min_{x\in\RR^d}\cL_{\lambda,Q}(x;b),
\]
any active algorithm that queries fewer than
\(
c\,\dlambda^\eps(A;Q)/\eps
\)
entries of \(b\) must fail on some response vector \(b\in\RR^{nd}\), in the sense that
\[
\PP_{\mathrm{alg}}\!\left(
\cL_{\lambda,Q}(\hat{x};b)
\le
\paren{1+c_0\eps}\opt_{\lambda,Q}(A,b)
\right)<0.75.
\]
\end{restatable}

The next corollary transfers this lower bound to a fixed ellipsoid whose radius is
chosen before the hard response vector.

\begin{restatable}{corollary}{effectiveballlowerboundcor}
\label{cor:effective-dimension-ball}
Under the setting of \Cref{thm:effective-dimension-lower-bound}, for a universal
constant \(c_0>0\), there is a radius \(R>0\) and an ellipsoid
\(\cC=\Set{x\in\RR^d}{x^\top Qx\le R}\) such that any randomized active algorithm
returning \(\hat{x}\in\cC\) after fewer than
\(c\,d_{\lambda R}^\eps(A;R^{-1}Q)/\eps\) queries must fail on some response vector
\(b\in\RR^{nd}\). More precisely, with
\(\opt_\cC\coloneqq\min_{x\in\cC}\norm{Ax-b}_2^2\),
\[
\PP_{\mathrm{alg}}\!\left(
\norm{A\hat{x}-b}_2^2
\le
\paren{1+c_0\eps}\opt_\cC
\right)<0.75.
\]
\end{restatable}

See Appendix~\ref{app:lower-bound-comparison} for a comparison with prior lower bounds.

\textbf{Remark on the untruncated regime.} For the hard instance in
\Cref{thm:effective-dimension-lower-bound}, the eigenvalues of \((\Id_d+\lambda Q)^{-1}\) are
\(\tau_i=(1+\lambda q_i)^{-1}\). If
\(\eps\le\min_i\tau_i=1/(1+\lambda\max_i q_i)\), then the truncation disappears:
\(\dlambda^\eps(A;Q)=\dlambda(A;Q)\). In this regime, the lower bound depends on
the full effective dimension and becomes \(\Omega(\dlambda(A;Q)/\eps)\).

This also matches the dimension term in the
convex upper bound of \Cref{thm:convex-ellipsoid-regularized} after the natural parameter
rescaling from the Lagrangian reduction: for
\(\cC=\Set{x}{x^\top Qx\le R}=\cE(R^{-1}Q)\), the relevant effective dimension from
\eqref{eq:constraint-induced-dimension}, evaluated at regularization parameter
\(\lambda R\), is
\(d_{\lambda R}(A;R^{-1}Q)=\dlambda(A;Q)\).
Hence, in the regime where the truncation disappears, the lower bound matches the \(\deff(\cC)/\eps\)-type dependence in \Cref{thm:convex-ellipsoid-regularized}.

For the construction in \Cref{thm:effective-dimension-lower-bound}, the loss separates
across coordinates. The proof relates sign errors in the retained coordinates to excess
loss, with weights \(\tau_i\). For general designs, a queried label may involve several
coordinates, and this argument need not give a sharp lower bound.

\section{Experiments}
\label{sec:experiments}

We test whether using the feasible set $\cC$ to select
labels reduces $\cL(\hat x)$ at a fixed sample size. We study
ellipsoidal constraints and then learn how a PDE's solution depends on
its parameters.

We compare ellipsoid-regularized leverage-score sampling with uniform,
leverage-score, and ridge leverage-score sampling. Each method draws $m$
rows independently with replacement and minimizes the reweighted squared
loss over $\cC$. Plots show $\cL(\hat x)/\opt_\cC$; repeated draws
reuse labels. \Cref{app:additional-experiments} gives the experimental
parameters and studies the choice of outer ellipsoid and $\lambda$.

\textbf{Ellipsoidal constraints.}
To study regression over an ellipsoidal feasible set, we construct an
instance with $n=3000$ and $d=60$. We give the ellipsoid six long axes and
54 short axes, so feasible coefficients can vary more in some directions
than in others. We construct $A$ so that $A^\top A=U\diag(s_1,\ldots,s_d)U^\top$, where $U$ is a random
orthonormal matrix and $s_j\in[0.8,1.2]$, so the design is well conditioned.
To make the ellipsoid axes differ from the eigenvectors of $A^\top A$,
we let $R$ rotate each coordinate pair $(j,j+6)$, $j\in[6]$,
by $35^\circ$ and define
\[
    Q=UR\diag(0.1\Id_6,10\Id_{54})R^\top U^\top,
    \qquad \cC=\cE(Q).
\]
Each rotation mixes distinct eigenspaces of $A^\top A$, so $Q$ and
$A^\top A$ do not commute.

\begin{figure}[tb]
    \centering
    \includegraphics[width=0.76\linewidth]{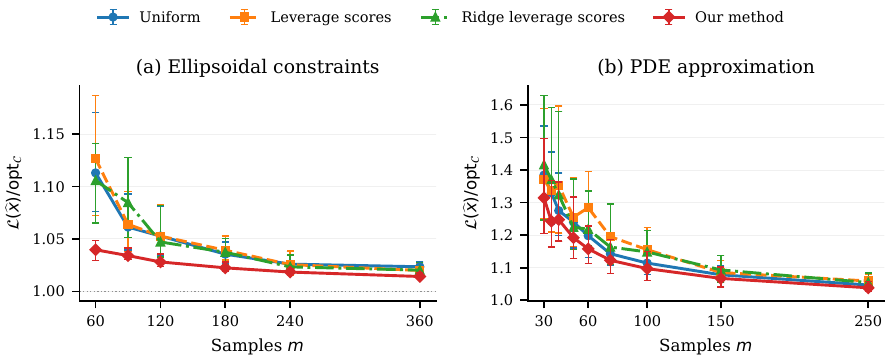}
    \caption{(a) Ellipsoidal constraints. (b) PDE approximation.
    Points and bars show medians and interquartile ranges of
    $\cL(\hat x)/\opt_\cC$ over 50 and 120 trials, respectively.}
    \label{fig:pde-rotated-experiments}
\end{figure}

We draw $\theta^\star$ uniformly from the sphere of radius $\sqrt{10}$
in the span of the ellipsoid's six longest axes. This places the signal
on the boundary where the constraint allows the greatest variation.
We generate $b=A\theta^\star+\xi$, with independent
$\xi_i\sim\mathcal N(0,0.1^2)$.
Across 50 trials with independent noise and row samples, our method has
the lowest median of $\cL(\hat x)/\opt_\cC$ at every tested sample size
(\Cref{fig:pde-rotated-experiments}(a)).
The advantage is greatest with few labels, despite the
different eigenspaces of $Q$ and $A^\top A$.

\textbf{Application to parametric PDEs.}
To study label selection when observations require solving a differential
equation, we consider steady heat conduction in a rod of unit length.
For conductivity parameters $\mu\in[-1,1]^3$, the temperature
$u(s;\mu)$ at position $s$ satisfies
\[
    -\frac{\mathrm d}{\mathrm ds}\left(\kappa(s;\mu)
    \frac{\mathrm du}{\mathrm ds}(s;\mu)\right)=1,
    \qquad s\in(0,1),\qquad u(0;\mu)=u(1;\mu)=0.
\]
The endpoints are held at zero temperature, the right-hand side represents
uniform heat generation, and $\kappa(s;\mu)>0$ is the thermal conductivity.
The three parameters determine its profile along the rod.

Our target $q(\mu)=u(1/2;\mu)$ is the midpoint temperature for each
conductivity profile. To predict it without a PDE solve for every parameter
choice, we fit $q$ by $\varphi(\mu)^\top x$. Smooth variation in
conductivity motivates a low-degree polynomial that models parameter
effects and interactions with few coefficients. We use $d=35$ known, normalized Legendre features
of total degree at most four; $x\in\RR^d$ contains the coefficients to learn.

To favor lower-degree terms, we impose
$\cC=\{x:|x_\alpha|\le r_{|\alpha|}\}$, where $|\alpha|$ is the total
degree of a feature and the chosen bounds $r_{|\alpha|}$ decrease with degree.
For $n=2000$ candidate parameter triples $\mu_i$, we set
$a_i=\varphi(\mu_i)/\sqrt n$ and
$b_i=(q(\mu_i)+\xi_i)/\sqrt n$, with independent
$\xi_i\sim\mathcal N(0,0.002^2)$. The matrix $A$ is known from the
parameter triples, whereas querying $b_i$ requires a numerical PDE solve.
Thus, learning the polynomial approximation reduces to the constrained
regression problem $\min_{x\in\cC}\cL(x)$ studied above. We seek
coefficients whose average squared prediction error over the parameter
pool is close to the constrained optimum, using few noisy PDE responses.

Our method uses the outer ellipsoid with shape matrix
$Q=\diag((d r_{|\alpha|}^2)^{-1})$ to incorporate the coefficient bounds
into its sampling probabilities, then fits $x$ over the original box $\cC$.
\Cref{app:experiment-pde-details} gives the construction details.

Over 120 trials with fixed labels and independently sampled rows, our
method again has the lowest median of $\cL(\hat x)/\opt_\cC$
(\Cref{fig:pde-rotated-experiments}(b)). At $m=35$, the median is $1.243$,
compared with $1.330$ for uniform sampling, the best baseline at this
sample size. The gap narrows as $m$ grows, so the greatest benefit occurs
when few PDE solves are available.

\section{Conclusion and future work}
\label{sec:conclusion}

This paper studied active constrained linear regression.
Our algorithm uses ellipsoid-regularized leverage scores to bias samples toward
directions that matter under an outer ellipsoidal approximation of the constraint set.
Our results show that incorporating constraint geometry into label selection can
substantially reduce the query complexity of constrained regression.
Our lower bounds show that the \(d/\varepsilon^2\) rate is unavoidable for general
non-convex constraints, and that the ellipsoid-regularized sampling mass must enter with
a \(1/\varepsilon\) factor for ellipsoidal constraints in a natural accuracy regime.

Several directions remain open.
First, outer ellipsoids may be too coarse for structured constraints such as sparse
sets, simplices, low-rank sets, and \(\ell_p\) balls; related sketching results for
sparse regression like \citet{MaiMunteanuMuscoRaoSchwiegelshohnWoodruff2023} suggest that
exploiting such structure directly may lead to sharper bounds.
Second, an important computational question is how to efficiently approximate the
outer ellipsoid minimizing the sampling mass for more general representations of
the feasible set.
Third, for general designs the sharp lower-bound measure may depend on more than the
regularized spectrum, including how the queryable rows expose the important directions
of the feasible set.
Finally, it remains open whether adaptive sampling, which uses earlier
responses to choose later queries, can improve the instance-dependent query bounds
achieved by non-adaptive methods.

\renewcommand{\acksection}{\section*{Acknowledgements}}
\providecommand{\additionalacknowledgments}{}
\begin{ack}
We thank Eric Price for suggesting the projection-based approach for convex
constraints\additionalacknowledgments.

CM is supported by NSF award 2045590.
SK is supported by an Amazon AI PhD fellowship.
YZ is supported by Institute for Foundations of Machine Learning (IFML).
\end{ack}

\bibliographystyle{arxiv_plainnat}
\bibliography{refs}

@techreport{Settles2009,
  author      = {Burr Settles},
  title       = {Active Learning Literature Survey},
  institution = {University of Wisconsin--Madison},
  type        = {Computer Sciences Technical Report},
  number      = {1648},
  year        = {2009}
}

@article{CohenDeVore2015,
  title={Approximation of high-dimensional parametric PDEs},
  author={Cohen, Albert and DeVore, Ronald},
  journal={Acta Numerica},
  volume={24},
  pages={1--159},
  year={2015},
  publisher={Cambridge University Press}
}

@book{AdcockBrugiapagliaWebster2022,
  title={Sparse polynomial approximation of high-dimensional functions},
  author={Adcock, Ben and Brugiapaglia, Simone and Webster, Clayton G},
  volume={25},
  year={2022},
  publisher={SIAM}
}

@article{ClarksonWoodruff2017,
  title={Low-rank approximation and regression in input sparsity time},
  author={Clarkson, Kenneth L and Woodruff, David P},
  journal={Journal of the ACM (JACM)},
  volume={63},
  number={6},
  pages={1--45},
  year={2017},
  publisher={ACM New York, NY, USA}
}

@article{PilanciWainwright2015,
  title={Randomized sketches of convex programs with sharp guarantees},
  author={Pilanci, Mert and Wainwright, Martin J},
  journal={IEEE Transactions on Information Theory},
  volume={61},
  number={9},
  pages={5096--5115},
  year={2015},
  publisher={IEEE}
}

@article{AlaouiMahoney2015,
  title={Fast randomized kernel ridge regression with statistical guarantees},
  author={Alaoui, Ahmed and Mahoney, Michael W},
  journal={Advances in neural information processing systems},
  volume={28},
  year={2015}
}

@article{AvronClarksonWoodruff2017,
  title={Sharper Bounds for Regularized Data Fitting},
  author={Avron, Haim and Clarkson, Kenneth L and Woodruff, David P},
  journal={Approximation, Randomization, and Combinatorial Optimization. Algorithms and Techniques},
  year={2017}
}

@inproceedings{CohenMuscoMusco2017,
  title={Input sparsity time low-rank approximation via ridge leverage score sampling},
  author={Cohen, Michael B and Musco, Cameron and Musco, Christopher},
  booktitle={Proceedings of the Twenty-Eighth Annual ACM-SIAM Symposium on Discrete Algorithms},
  pages={1758--1777},
  year={2017},
  organization={SIAM}
}

@article{DerezinskiWarmuthHsu2018,
  title={Leveraged volume sampling for linear regression},
  author={Derezinski, Michal and Warmuth, Manfred K. and Hsu, Daniel J},
  journal={Advances in Neural Information Processing Systems},
  volume={31},
  year={2018}
}

@inproceedings{TangGolbabaeeDavies2017,
  title={Gradient projection iterative sketch for large-scale constrained least-squares},
  author={Tang, Junqi and Golbabaee, Mohammad and Davies, Mike E},
  booktitle={International Conference on Machine Learning},
  pages={3377--3386},
  year={2017},
  organization={PMLR}
}

@inproceedings{MaiMunteanuMuscoRaoSchwiegelshohnWoodruff2023,
  title={Optimal sketching bounds for sparse linear regression},
  author={Mai, Tung and Munteanu, Alexander and Musco, Cameron and Rao, Anup and Schwiegelshohn, Chris and Woodruff, David},
  booktitle={International Conference on Artificial Intelligence and Statistics},
  pages={11288--11316},
  year={2023},
  organization={PMLR}
}

@inproceedings{ChenPrice:2019a,
  title={Active regression via linear-sample sparsification},
  author={Chen, Xue and Price, Eric},
  booktitle={Conference on Learning Theory},
  pages={663--695},
  year={2019},
  organization={PMLR}
}

@article{rajaraman2022semi,
  title={Semi-supervised active linear regression},
  author={Devvrit, Fnu and Rajaraman, Nived and Awasthi, Pranjal},
  journal={Advances in Neural Information Processing Systems},
  volume={35},
  pages={1294--1306},
  year={2022}
}

@inproceedings{ShimizuChengMusco:2024,
  title={Improved active learning via dependent leverage score sampling},
  author={Shimizu, Atsushi and Cheng, Xiaoou and Musco, Christopher and Weare, Jonathan},
  booktitle={The Twelfth International Conference on Learning Representations},
  year={2024},
  url={https://proceedings.iclr.cc/paper_files/paper/2024/hash/7f05193e5487287a890df7fbc3554427-Abstract-Conference.html}
}

@book{lattimore2020bandit,
  title={Bandit algorithms},
  author={Lattimore, Tor and Szepesv{\'a}ri, Csaba},
  year={2020},
  publisher={Cambridge University Press}
}

@article{LaurentMassart2000,
  title={Adaptive estimation of a quadratic functional by model selection},
  author={Laurent, Beatrice and Massart, Pascal},
  journal={Annals of statistics},
  pages={1302--1338},
  year={2000},
  publisher={JSTOR}
}

@article{WangParkSusztakZhangLi2019,
  title={Bulk tissue cell type deconvolution with multi-subject single-cell expression reference},
  author={Wang, Xuran and Park, Jihwan and Susztak, Katalin and Zhang, Nancy R and Li, Mingyao},
  journal={Nature communications},
  volume={10},
  number={1},
  pages={380},
  year={2019},
  publisher={Nature Publishing Group UK London}
}

@article{HeinzChang2001,
  title={Fully constrained least squares linear spectral mixture analysis method for material quantification in hyperspectral imagery},
  author={Heinz, Daniel C and Chang, Chein-I},
  journal={IEEE transactions on geoscience and remote sensing},
  volume={39},
  number={3},
  pages={529--545},
  year={2001},
  publisher={IEEE}
}

@article{AbadieDiamondHainmueller2010,
  title={Synthetic control methods for comparative case studies: Estimating the effect of California’s tobacco control program},
  author={Abadie, Alberto and Diamond, Alexis and Hainmueller, Jens},
  journal={Journal of the American statistical Association},
  volume={105},
  number={490},
  pages={493--505},
  year={2010},
  publisher={Taylor \& Francis}
}

@article{WuYangLiu2014,
  title={Nonnegative-lasso and application in index tracking},
  author={Wu, Lan and Yang, Yuehan and Liu, Hanzhong},
  journal={Computational Statistics \& Data Analysis},
  volume={70},
  pages={116--126},
  year={2014},
  publisher={Elsevier}
}

@article{LeeBattleRainaNg2006,
  title={Efficient sparse coding algorithms},
  author={Lee, Honglak and Battle, Alexis and Raina, Rajat and Ng, Andrew},
  journal={Advances in neural information processing systems},
  volume={19},
  year={2006}
}

@article{Chatterjee2014,
  title={A NEW PERSPECTIVE ON LEAST SQUARES UNDER CONVEX CONSTRAINT},
  author={Chatterjee, Sourav},
  journal={The Annals of Statistics},
  pages={2340--2381},
  year={2014},
  publisher={JSTOR}
}

@article{ChandrasekaranRechtParriloWillsky2012,
  title={The convex geometry of linear inverse problems},
  author={Chandrasekaran, Venkat and Recht, Benjamin and Parrilo, Pablo A and Willsky, Alan S},
  journal={Foundations of Computational mathematics},
  volume={12},
  number={6},
  pages={805--849},
  year={2012},
  publisher={Springer}
}

@article{AmelunxenLotzMcCoyTropp2014,
  title={Living on the edge: Phase transitions in convex programs with random data},
  author={Amelunxen, Dennis and Lotz, Martin and McCoy, Michael B and Tropp, Joel A},
  journal={Information and Inference: A Journal of the IMA},
  volume={3},
  number={3},
  pages={224--294},
  year={2014},
  publisher={OUP}
}

@article{NegahbanRavikumarWainwrightYu2012,
  title={A Unified Framework for High-Dimensional Analysis of M-Estimators with Decomposable Regularizers},
  author={Negahban, Sahand N and Ravikumar, Pradeep and Wainwright, Martin J and Yu, Bin},
  journal={Statistical Science},
  volume={27},
  number={4},
  pages={538--557},
  year={2012}
}

@article{PilanciWainwright2016,
  title={Iterative Hessian sketch: Fast and accurate solution approximation for constrained least-squares},
  author={Pilanci, Mert and Wainwright, Martin J},
  journal={Journal of Machine Learning Research},
  volume={17},
  number={53},
  pages={1--38},
  year={2016}
}

@inproceedings{YangZhangChenZhangChen2017,
  title={Suggestive annotation: A deep active learning framework for biomedical image segmentation},
  author={Yang, Lin and Zhang, Yizhe and Chen, Jianxu and Zhang, Siyuan and Chen, Danny Z},
  booktitle={International conference on medical image computing and computer-assisted intervention},
  pages={399--407},
  year={2017},
  organization={Springer}
}

@inproceedings{Sarlos2006,
  title={Improved approximation algorithms for large matrices via random projections},
  author={Sarlos, Tamas},
  booktitle={2006 47th annual IEEE symposium on foundations of computer science (FOCS'06)},
  pages={143--152},
  year={2006},
  organization={IEEE}
}

@article{WarmuthLiaoRaetschMathiesonPuttaLemmen2003,
  title={Active learning with support vector machines in the drug discovery process},
  author={Warmuth, Manfred K and Liao, Jun and R{\"a}tsch, Gunnar and Mathieson, Michael and Putta, Santosh and Lemmen, Christian},
  journal={Journal of chemical information and computer sciences},
  volume={43},
  number={2},
  pages={667--673},
  year={2003},
  publisher={ACS Publications}
}

@article{LookmanBalachandranXueYuan2019,
  title={Active learning in materials science with emphasis on adaptive sampling using uncertainties for targeted design},
  author={Lookman, Turab and Balachandran, Prasanna V and Xue, Dezhen and Yuan, Ruihao},
  journal={npj Computational Materials},
  volume={5},
  number={1},
  pages={21},
  year={2019},
  publisher={Nature Publishing Group UK London}
}

@article{ShieldsGurleyCatarelliChauhanOjedaTuzMasters2023,
  title={Active learning applied to automated physical systems increases the rate of discovery},
  author={Shields, Michael D and Gurley, Kurtis and Catarelli, Ryan and Chauhan, Mohit and Ojeda-Tuz, Mariel and Masters, Forrest J},
  journal={Scientific Reports},
  volume={13},
  number={1},
  pages={8402},
  year={2023},
  publisher={Nature Publishing Group UK London}
}

@article{DrineasKannanMahoney2006,
  author = {Drineas, Petros and Kannan, Ravi and Mahoney, Michael W.},
  title = {Fast Monte Carlo Algorithms for Matrices I: Approximating Matrix Multiplication},
  journal = {SIAM Journal on Computing},
  volume = {36},
  number = {1},
  pages = {132--157},
  year = {2006},
  doi = {10.1137/S0097539704442684}
}

@InProceedings{pmlr-v247-gajjar24a,
  title = 	 {Agnostic Active Learning of Single Index Models with Linear Sample Complexity},
  author =       {Gajjar, Aarshvi and Tai, Wai Ming and Xingyu, Xu and Hegde, Chinmay and Musco, Christopher and Li, Yi},
  booktitle = 	 {Proceedings of Thirty Seventh Conference on Learning Theory},
  pages = 	 {1715--1754},
  year = 	 {2024},
  volume = 	 {247},
  publisher =    {PMLR},
}

@InProceedings{pmlr-v206-gajjar23a,
  title = 	 {Active Learning for Single Neuron Models with Lipschitz Non-Linearities},
  author =       {Gajjar, Aarshvi and Musco, Christopher and Hegde, Chinmay},
  booktitle = 	 {Proceedings of The 26th International Conference on Artificial Intelligence and Statistics},
  pages = 	 {4101--4113},
  year = 	 {2023},
  volume = 	 {206},
  publisher =    {PMLR},
}

@article{Ozsar_2025,
doi = {10.1088/1361-6420/ada798},
year = {2025},
publisher = {IOP Publishing},
volume = {41},
number = {2},
pages = {025004},
author = {Ozsar, Ege and Kilmer, Misha and de Sturler, Eric and Saibaba, Arvind K and Miller, Eric},
title = {Parametric level-sets enhanced to improve reconstruction (PaLEnTIR)},
journal = {Inverse Problems},
}

@inproceedings{ChenDerezinski2021,
  title={Query complexity of least absolute deviation regression via robust uniform convergence},
  author={Chen, Xue and Derezinski, Michal},
  booktitle={Conference on Learning Theory},
  pages={1144--1179},
  year={2021},
  organization={PMLR}
}

@article{DevvritRajaramanAwasthi2021v1,
  title={Semi-supervised Active Regression},
  author={Devvrit, Fnu and Rajaraman, Nived and Awasthi, Pranjal},
  journal={arXiv preprint arXiv:2106.06676},
  year={2021},
  note={Version 1, June 12, 2021}
}

@inproceedings{CohenCousinsLeeYang2019,
  author    = {Cohen, Michael B. and Cousins, Ben and Lee, Yin Tat and Yang, Xin},
  title     = {A near-optimal algorithm for approximating the {John Ellipsoid}},
  booktitle = {Proceedings of the Thirty-Second Conference on Learning Theory},
  series    = {Proceedings of Machine Learning Research},
  volume    = {99},
  pages     = {849--873},
  year      = {2019},
  publisher = {PMLR},
  url       = {https://proceedings.mlr.press/v99/cohen19a.html}
}

@inproceedings{WoodruffYasuda2024,
  author    = {Woodruff, David P. and Yasuda, Taisuke},
  title     = {{John Ellipsoids} via Lazy Updates},
  booktitle = {Advances in Neural Information Processing Systems},
  volume    = {37},
  year      = {2024},
  doi       = {10.52202/079017-2179}
}

@article{GulerGurtuna2012,
  author  = {G{\"u}ler, Osman and G{\"u}rtuna, Filiz},
  title   = {Symmetry of convex sets and its applications to the extremal ellipsoids of convex bodies},
  journal = {Optimization Methods and Software},
  volume  = {27},
  number  = {4--5},
  pages   = {735--759},
  year    = {2012},
  doi     = {10.1080/10556788.2011.626037}
}

\appendix

\clearpage
\section{Proofs for constraint-aware sampling}
\label{app:constraint-aware-proofs}

We first restate a result from \citep{AlaouiMahoney2015} for approximating matrix products.

\begin{theorem}[Theorem~2 in \citep{AlaouiMahoney2015}]
\label{thm:alaoui-mahoney-concentration}
Let \(\Psi\in\RR^{q\times N}\) be a matrix with columns
\(\psi_1,\ldots,\psi_N\), and suppose \(\norm{\Psi}_F>0\). Suppose we
sample columns independently with replacement using \(
p_i=\frac{\norm{\psi_i}_2^2}{\norm{\Psi}_F^2},
\)
for \(i\in[N]\), and let \(S\in\RR^{m\times N}\) be the sampling-and-rescaling matrix from
\eqref{eq:sampling-matrix}. Then for every \(t>0\),
\[
\PP\!\paren{
\lambda_{\max}\!\paren{\Psi\Psi^\top-\Psi S^\top S\Psi^\top}\ge t
}
\le
q\exp\!\paren{
-\frac{m t^2/2}
{\lambda_{\max}(\Psi\Psi^\top)\paren{\norm{\Psi}_F^2+t/3}}
},
\]

where \(\lambda_{\max}(T)\) is the maximum eigen value of a matrix \(T\).
\end{theorem}

The next lemma follows almost directly from
\Cref{thm:alaoui-mahoney-concentration}, a restatement of
Theorem~2 of \citet{AlaouiMahoney2015}; the proof below chooses
\(\Psi\) so that row sampling for \(A\) becomes column sampling for \(\Psi\).

\begin{lemma}[One-sided subspace embedding]
\label{lem:one-sided-ridge-subspace-embedding}
Let \(A\in\RR^{n\times d}\), \(Q\succ0\), and \(\lambda\ge0\) satisfy
\(A^\top A+\lambda Q\succ0\). With \(\lev_i(A;\lambda,Q)\) as in
\eqref{eq:ellipsoid-lev-score} and \(\dlambda(A;Q)\) as in
\eqref{eq:regularized-score-mass}. Form the
sampling-and-rescaling matrix \(S\in\RR^{m\times n}\) from
\eqref{eq:sampling-matrix} with probabilities
\(
p_i=\frac{\lev_i(A;\lambda,Q)}{\dlambda(A;Q)}
\qquad (i\in[n]).
\)
Then for any \(\eps,\delta\in(0,1)\), if \( m \ge C\,\frac{\dlambda(A;Q)}{\eps^2}\log\!\paren{\frac{d}{\delta}}
\) for a sufficiently large constant \(C > 0\), with probability at least \(1-\delta\),
\begin{equation}
\label{eq:one-sided-ridge-subspace-embedding}
\norm{SAx}_2^2+\lambda\qnorm{x}{Q}^2
\ge
(1-\eps)\paren{\norm{Ax}_2^2+\lambda\qnorm{x}{Q}^2}
\qquad \text{for all } x\in\RR^d.
\end{equation}
\end{lemma}

\begin{proof}
Let \(H \coloneqq A^\top A+\lambda Q\in\RR^{d\times d}\), and set
\(\Psi\coloneqq H^{-1/2}A^\top\in\RR^{d\times n}\). The \(i\)-th column of \(\Psi\) is
\(\psi_i=H^{-1/2}A^\top e_i\), implying that
\(
\norm{\psi_i}_2^2
=
e_i^\top AH^{-1}A^\top e_i
=
\lev_i(A;\lambda,Q).
\)

Further, the Frobenius norm of \(\Psi\) is the total regularized leverage score mass:
\[
\norm{\Psi}_F^2=\sum_{i=1}^n\norm{\psi_i}_2^2=\dlambda(A;Q)
\]

Since \(A^\top A\preceq H\), 
\(
\Psi\Psi^\top=H^{-1/2}A^\top AH^{-1/2}\preceq I
\).

Using these two results, we now show the following Löwner ordering holds.
\begin{equation}
\label{eq:sampled-gram-one-sided}
\Psi S^\top S\Psi^\top \succeq \Psi\Psi^\top-\eps I.
\end{equation}
Let us consider two cases. First, when \(\dlambda(A;Q)\le\eps\). Here, we have that \(\Psi\Psi^\top\preceq \norm{\Psi}_F^2I\preceq\eps I\), so
\eqref{eq:sampled-gram-one-sided} follows deterministically. 

Second, when \(\dlambda(A;Q)>\eps\). In this case we can invoke \Cref{thm:alaoui-mahoney-concentration} for \(p_i=\norm{\psi_i}_2^2/\norm{\Psi}_F^2\). We obtain the following result.
\[
\PP\!\paren{\lambda_{\max}\!\paren{\Psi\Psi^\top-\Psi S^\top S\Psi^\top}\ge\eps}
\le
d\exp\!\paren{
-\frac{m\eps^2/2}
{\lambda_{\max}(\Psi\Psi^\top)\paren{\dlambda(A;Q)+\eps/3}}
}.
\]
Since \(\Psi\Psi^\top\preceq I\), we have
\(\lambda_{\max}(\Psi\Psi^\top)\le1\). Since \(\dlambda(A;Q)>\eps\),
\[
\dlambda(A;Q)+\eps/3 \le \frac43\dlambda(A;Q).
\]
Thus, for a universal constant \(c>0\), the preceding probability is at most
\[
d\exp\!\paren{-c\,\frac{m\eps^2}{\dlambda(A;Q)}}.
\]
Taking \(m\ge C\dlambda(A;Q)\eps^{-2}\log(d/\delta)\) with \(C\) large
enough makes this failure probability at most \(\delta\).
Therefore, with probability at least \(1-\delta\),
\[
\lambda_{\max}\!\paren{\Psi\Psi^\top-\Psi S^\top S\Psi^\top}<\eps,
\]
which implies
\(\Psi\Psi^\top-\Psi S^\top S\Psi^\top\preceq \eps I\), or equivalently
\eqref{eq:sampled-gram-one-sided}.

We now derive the main statement. For any \(x\in\RR^d\),
\[
\norm{SAx}_2^2
=
x^\top H^{1/2}\Psi S^\top S\Psi^\top H^{1/2}x.
\]
Further,
\[
\lambda\qnorm{x}{Q}^2
=
x^\top H^{1/2}(I-\Psi\Psi^\top)H^{1/2}x,
\]
because \(H^{1/2}(I-\Psi\Psi^\top)H^{1/2}=H-A^\top A=\lambda Q\).
From \eqref{eq:sampled-gram-one-sided},
\[
\norm{SAx}_2^2+\lambda\qnorm{x}{Q}^2
=
x^\top H^{1/2}\paren{\Psi S^\top S\Psi^\top+I-\Psi\Psi^\top}H^{1/2}x
\ge
(1-\eps)x^\top Hx.
\]
Finally, \(x^\top Hx=\norm{Ax}_2^2+\lambda\qnorm{x}{Q}^2\), which gives us the main statement.
\end{proof}

The next lemma also follows directly from an approximate matrix multiplication (AMM) analysis, \citet[Section~4]{DrineasKannanMahoney2006}, which is standard for active learning using leverage score sampling.

\begin{lemma}[Ridge Approx Matrix Multiplication]
\label{lem:ridge-amm-residual-cross-term}
Let \(H \coloneqq A^\top A+\lambda Q\), where \(Q\succ0\), and assume \(H\succ0\). With \(\lev_i(A;\lambda,Q)\) as in \eqref{eq:ellipsoid-lev-score} and \(\dlambda(A;Q)\) as in \eqref{eq:regularized-score-mass}, suppose \(\dlambda(A;Q)>0\).

If we sample rows independently with replacement using \(p_i=\lev_i(A;\lambda,Q)/\dlambda(A;Q)\), and let \(S\in\RR^{m\times n}\) be the sample-and-rescale matrix from \eqref{eq:sampling-matrix}, then for any fixed vector \(r\in\RR^n\) and any \(\eps,\delta\in(0,1)\), if \(m\ge \dlambda(A;Q)/(\eps^2\delta)\), with probability at least \(1-\delta\),
\begin{equation}
\label{eq:ridge-amm-cross-term}
\abs{x^\top A^\top(S^\top S-I)r}
\le
\eps\sqrt{x^\top Hx}\,\norm{r}_2
\qquad \text{for all } x\in\RR^d.
\end{equation}
\end{lemma}

\begin{proof}
Set \(\Psi\coloneqq H^{-1/2}A^\top\in\RR^{d\times n}\), and write
\(\psi_i\) for its \(i\)-th column. Then
\[
\norm{\psi_i}_2^2
=
(AH^{-1}A^\top)_{ii}
=
\lev_i(A;\lambda,Q).
\]
Hence \(p_i=\norm{\psi_i}_2^2/\dlambda(A;Q)\). Rows with
\(\norm{\psi_i}_2=0\) have \(p_i=0\) and contribute neither to the sampling nor
to the sums below, so all displayed ratios are over indices with \(p_i>0\).

Set
\[
g\coloneqq H^{-1/2}A^\top(S^\top S-I)r
=
\Psi(S^\top S-I)r.
\]
If \(j_1,\ldots,j_m\) are the sampled indices, then
\[
S^\top S
=
\frac1m\sum_{t=1}^m \frac{1}{p_{j_t}}e_{j_t}e_{j_t}^\top.
\]
Consequently,
\begin{align*}
\Psi S^\top Sr
&=
\Psi\paren{\frac1m\sum_{t=1}^m
\frac{1}{p_{j_t}}e_{j_t}e_{j_t}^\top}r \\
&=
\frac1m\sum_{t=1}^m
\frac{1}{p_{j_t}}\Psi e_{j_t}\,e_{j_t}^\top r \\
&=
\frac1m\sum_{t=1}^m
\frac{\psi_{j_t}r_{j_t}}{p_{j_t}}.
\end{align*}
For a single draw \(j\sim p\), the summand is unbiased:
\[
\EE\!\bracket{\frac{\psi_jr_j}{p_j}}
=
\sum_{i:\,p_i>0} p_i\frac{\psi_ir_i}{p_i}
=
\sum_{i:\,p_i>0} \psi_ir_i
=
\sum_{i=1}^n \psi_ir_i
=
\Psi r.
\]
Thus \(g\) is the average of \(m\) independent centered copies of
\(\psi_jr_j/p_j-\Psi r\). Independence gives
\begin{align*}
\EE[\norm{g}_2^2]
&=
\frac1m
\EE\!\bracket{\norm{\frac{\psi_jr_j}{p_j}-\Psi r}_2^2} \\
&\le
\frac1m
\EE\!\bracket{\norm{\frac{\psi_jr_j}{p_j}}_2^2} \\
&=
\frac1m\sum_{i:\,p_i>0} p_i
\norm{\frac{\psi_ir_i}{p_i}}_2^2 \\
&=
\frac1m\sum_{i:\,p_i>0}
\frac{\norm{\psi_i}_2^2}{p_i}r_i^2 \\
&\le
\frac{\dlambda(A;Q)}{m}\norm{r}_2^2,
\end{align*}
where we used \(p_i=\norm{\psi_i}_2^2/\dlambda(A;Q)\).

If \(r=0\), the claimed bound is immediate. Otherwise, Markov's inequality gives
\[
\PP\!\paren{\norm{g}_2>\eps\norm{r}_2}
\le
\frac{\EE[\norm{g}_2^2]}{\eps^2\norm{r}_2^2}
\le
\frac{\dlambda(A;Q)}{m\eps^2}.
\]
If \(m\ge \dlambda(A;Q)/(\eps^2\delta)\), this probability is at most \(\delta\),
so with probability at least \(1-\delta\), \(\norm{g}_2\le\eps\norm{r}_2\).

Finally, for any \(x\in\RR^d\),
\[
x^\top A^\top(S^\top S-I)r
=
(H^{1/2}x)^\top \Psi(S^\top S-I)r.
\]
On this event, the right-hand side is
\((H^{1/2}x)^\top g\). Therefore Cauchy--Schwarz gives
\[
\abs{x^\top A^\top(S^\top S-I)r}
\le
\norm{H^{1/2}x}_2\norm{g}_2
\le
\eps\sqrt{x^\top Hx}\,\norm{r}_2,
\]
which proves the lemma.
\end{proof}

\generalconstraintawarethm*

\begin{proof}[Proof of \Cref{thm:general-constraint-aware}]
Let
\[
x^\star \in \argmin_{u\in\cC}\norm{Au-b}_2^2,
\qquad
r\coloneqq Ax^\star-b,
\qquad
x\coloneqq \hat{x}-x^\star,
\]
so that \(\norm{r}_2^2=\opt_\cC\). We also write
\(\lambda=\lambda_\star\) and \(H=A^\top A+\lambda Q\). By the choice of
\(Q\) in the optimization
problem \eqref{eq:constraint-induced-dimension}, and by setting
\(\lambda=\lambda_\star\le c_\lambda\frac{\eps}{1+\eps}\opt_\cC
=O(\eps\opt_\cC)\), we have
\(\dlambda(A;Q)\le2\deff(\cC)\).

Let \(c>0\) be a sufficiently small absolute constant. Apply
\Cref{lem:one-sided-ridge-subspace-embedding} and
\Cref{lem:ridge-amm-residual-cross-term} with accuracy parameter \(c\eps\),
failure probability \(\delta/2\), and fixed vector \(r\) in the second lemma.
The stated lower bound on \(m\), with a sufficiently large absolute constant,
guarantees both events simultaneously with probability at least \(1-\delta\).
We now condition on these two events.

The optimality of \(\hat{x}\) for the sampled constrained problem gives
\[
\norm{S(A\hat{x}-b)}_2^2
\le
\norm{S(Ax^\star-b)}_2^2.
\]
Since \(A\hat{x}-b=Ax+r\), this is equivalent to
\begin{equation}
\label{eq:sampled-optimality-expanded}
\norm{SAx}_2^2+2x^\top A^\top S^\top S r\le0.
\end{equation}
The true excess loss is
\[
G
\coloneqq
\norm{A\hat{x}-b}_2^2-\norm{Ax^\star-b}_2^2
=
\norm{Ax}_2^2+2x^\top A^\top r.
\]
Combining this identity with \eqref{eq:sampled-optimality-expanded} gives
\begin{align}
G
&\le
\norm{Ax}_2^2-\norm{SAx}_2^2
+2x^\top A^\top(I-S^\top S)r \notag \\
&\le
\norm{Ax}_2^2-\norm{SAx}_2^2
+2\abs{x^\top A^\top(S^\top S-I)r}.
\label{eq:excess-basic-bound}
\end{align}

The one-sided embedding event implies
\[
\norm{Ax}_2^2-\norm{SAx}_2^2
\le
c\eps\paren{\norm{Ax}_2^2+\lambda\qnorm{x}{Q}^2}
=
c\eps\, x^\top Hx.
\]
The approximate matrix multiplication event implies
\[
\abs{x^\top A^\top(S^\top S-I)r}
\le
c\eps\sqrt{x^\top Hx}\,\norm{r}_2
=
c\eps\sqrt{x^\top Hx}\,\sqrt{\opt_\cC}.
\]
Substituting these two bounds into \eqref{eq:excess-basic-bound}, and then
using
\(2\sqrt{x^\top Hx\,\opt_\cC}\le x^\top Hx+\opt_\cC\), yields
\begin{equation}
\label{eq:excess-ridge-norm-bound}
G
\le
2c\eps\, x^\top Hx+c\eps\opt_\cC.
\end{equation}

It remains to control \(x^\top Hx\) using feasibility. Since
\(\cC\subseteq\cE(Q)\), both \(x^\star\) and \(\hat{x}\) satisfy
\(\qnorm{x^\star}{Q}\le1\) and \(\qnorm{\hat{x}}{Q}\le1\). Hence
\(\qnorm{x}{Q}\le2\). Also, because \(G\ge0\),
\[
\norm{Ax}_2^2
\le
\paren{\norm{A\hat{x}-b}_2+\norm{Ax^\star-b}_2}^2
=
\paren{\sqrt{\opt_\cC+G}+\sqrt{\opt_\cC}}^2
\le
2G+4\opt_\cC.
\]
Therefore
\begin{equation}
\label{eq:ridge-norm-feasibility-bound}
x^\top Hx
=
\norm{Ax}_2^2+\lambda\qnorm{x}{Q}^2
\le
2G+4\opt_\cC+4\lambda.
\end{equation}
Combining \eqref{eq:excess-ridge-norm-bound} and
\eqref{eq:ridge-norm-feasibility-bound} gives
\[
G
\le
4c\eps G
+9c\eps\opt_\cC
+8c\eps\lambda.
\]
Choose \(c\) sufficiently small so that \(4c\eps\le1/2\) for all
\(\eps\in(0,1/2)\). If \(\opt_\cC=0\), then
\(\lambda=0\), and the preceding display gives \(G\le4c\eps G\), hence
\(G=0\). Otherwise, since
\(\lambda=\lambda_\star\le c_\lambda\frac{\eps}{1+\eps}\opt_\cC\), the
preceding display implies
\[
G
\le
2c(9+8c_\lambda)\eps\opt_\cC.
\]
Taking \(c\) small enough, and absorbing the corresponding factor into the
absolute constant in the sample size, gives \(G\le\eps\opt_\cC\).
Thus
\[
\cL(\hat{x})
=
\opt_\cC+G
\le
(1+\eps)\opt_\cC.
\]
\end{proof}

\section{Proofs for improved convex-constraint bounds}
\label{app:convex-projection-proofs}

\begin{lemma}[Auxiliary projection fact]
\label{lem:proj-ineq}
Let \(K \subseteq \RR^n\) be closed and convex, let \(P_K\) denote Euclidean
projection onto \(K\), and set \(y=P_K(y_u)\) and \(z=P_K(z_u)\). Then
\begin{equation}
\label{eq:proj-ineq}
\norm{y_u-z}_2^2 - \norm{y_u-y}_2^2
\le
\norm{y_u-z_u}_2^2.
\end{equation}
\end{lemma}

\begin{proof}[Proof of \Cref{lem:proj-ineq}]
Since \(y,z \in K\) and \(z = P_K(z_u)\), the projection optimality condition gives
\[
\inner{z_u-z}{y-z} \le 0.
\]
Therefore,
\[
\begin{aligned}
\norm{y_u-z_u}_2^2
&=
\norm{(y-z) + \paren{(y_u-y)-(z_u-z)}}_2^2 \\
&=
\norm{y-z}_2^2
+
\norm{(y_u-y)-(z_u-z)}_2^2 + 2\inner{y-z}{y_u-y} - 2\inner{y-z}{z_u-z} \\
&\ge
\norm{y-z}_2^2 + 2\inner{y-z}{y_u-y}.
\end{aligned}
\]
Expanding \(\norm{(y_u-y)+(y-z)}_2^2 = \norm{y_u-z}_2^2\) shows that the final line equals
\[
\norm{y_u-z}_2^2 - \norm{y_u-y}_2^2,
\]
which proves \eqref{eq:proj-ineq}.
\end{proof}

\begin{lemma}
\label{lem:deterministic-projection-reduction}
Let \(\cC \subseteq \RR^d\) be nonempty, closed, and convex. For any
\(\hat{x}_{\mathrm{u}}\), let
\(\hat{x}_{\mathrm{proj}}\in\argmin_{x \in \cC}
\allowbreak\norm{A\hat{x}_{\mathrm{u}}-Ax}_2^2\), and let
\(x_{\mathrm{u}}^\star\in\argmin_{x \in \RR^d}\norm{Ax-b}_2^2\). Then
\[
\norm{A\hat{x}_{\mathrm{proj}}-b}_2^2-\opt_\cC
\le
\norm{A\hat{x}_{\mathrm{u}}-b}_2^2
-
\norm{Ax_{\mathrm{u}}^\star-b}_2^2.
\]
\end{lemma}

\begin{proof}[Proof of \Cref{lem:deterministic-projection-reduction}]
Let
\[
r_{\mathrm{u}} \coloneqq b-Ax_{\mathrm{u}}^\star.
\]
Because \(x_{\mathrm{u}}^\star\) is an unconstrained least-squares minimizer,
\(r_{\mathrm{u}}\) is orthogonal to \(\col(A)\). Hence for every
\(y \in \col(A)\),
\begin{equation}
\label{eq:convex-projection-pythagorean}
\norm{y-b}_2^2
=
\norm{r_{\mathrm{u}}}_2^2 + \norm{y-Ax_{\mathrm{u}}^\star}_2^2.
\end{equation}

Set
\[
K \coloneqq A\cC.
\]
By the standing assumption on \(A\), the map \(x \mapsto Ax\) is a
homeomorphism from \(\RR^d\) onto \(\col(A)\). Thus \(K\) is closed and convex.
Let \(x_\cC^\star\in\argmin_{x \in \cC}\norm{Ax-b}_2^2\).
By \eqref{eq:convex-projection-pythagorean}, minimizing
\(\norm{y-b}_2^2\) over \(y \in A\cC\) is equivalent to minimizing
\(\norm{y-Ax_{\mathrm{u}}^\star}_2^2\) over \(y \in K\). Therefore
\(Ax_\cC^\star\) is a metric projection of \(Ax_{\mathrm{u}}^\star\) onto
\(K\). Likewise, \(A\hat{x}_{\mathrm{proj}}\) is a metric projection of
\(A\hat{x}_{\mathrm{u}}\) onto \(K\).

Applying \Cref{lem:proj-ineq} with
\[
y_u \coloneqq Ax_{\mathrm{u}}^\star,
\qquad
z_u \coloneqq A\hat{x}_{\mathrm{u}},
\qquad
y \coloneqq Ax_\cC^\star,
\qquad
z \coloneqq A\hat{x}_{\mathrm{proj}},
\]
gives
\begin{equation}
\label{eq:convex-projection-gap}
\norm{Ax_{\mathrm{u}}^\star-A\hat{x}_{\mathrm{proj}}}_2^2
-
\norm{Ax_{\mathrm{u}}^\star-Ax_\cC^\star}_2^2
\le
\norm{A\hat{x}_{\mathrm{u}}-Ax_{\mathrm{u}}^\star}_2^2.
\end{equation}
Applying \eqref{eq:convex-projection-pythagorean} with
\(y = A\hat{x}_{\mathrm{proj}}\) and \(y = Ax_\cC^\star\), and subtracting
the two resulting identities, gives
\begin{equation}
\label{eq:convex-excess-risk-identity}
\norm{A\hat{x}_{\mathrm{proj}}-b}_2^2
-
\norm{Ax_\cC^\star-b}_2^2
=
\norm{Ax_{\mathrm{u}}^\star-A\hat{x}_{\mathrm{proj}}}_2^2
-
\norm{Ax_{\mathrm{u}}^\star-Ax_\cC^\star}_2^2.
\end{equation}
Combining \eqref{eq:convex-projection-gap} and
\eqref{eq:convex-excess-risk-identity} yields
\[
\norm{A\hat{x}_{\mathrm{proj}}-b}_2^2
-
\norm{Ax_\cC^\star-b}_2^2
\le
\norm{A\hat{x}_{\mathrm{u}}-Ax_{\mathrm{u}}^\star}_2^2.
\]
Finally, the orthogonality of \(r_{\mathrm{u}}\) gives
\[
\norm{A\hat{x}_{\mathrm{u}}-b}_2^2
-
\norm{Ax_{\mathrm{u}}^\star-b}_2^2
=
\norm{A\hat{x}_{\mathrm{u}}-Ax_{\mathrm{u}}^\star}_2^2.
\]
This proves the additive excess-risk inequality.
\end{proof}

We now restate the theorem statements before giving their proofs.

The first statement is the convex projection guarantee obtained using the
standard constant-probability leverage-score sampling guarantee.

\convexprojectionthm*

\begin{proof}[Proof of \Cref{thm:convex-projection}]
By the unconstrained leverage-score sampling guarantee of
\citet[Claim~13]{Sarlos2006}, applied with accuracy parameter
\(\epsilon/3\), the stated lower bound on \(m\) implies that, with fixed
nonzero probability,
\[
\norm{A\hat{x}_{\mathrm{u}}-b}_2^2
-
\min_{x\in\RR^d}\norm{Ax-b}_2^2
\le
\eps\min_{x\in\RR^d}\norm{Ax-b}_2^2.
\]
Applying \Cref{lem:deterministic-projection-reduction} gives
\[
\norm{A\hat{x}_{\mathrm{proj}}-b}_2^2-\opt_\cC
\le
\eps\min_{x\in\RR^d}\norm{Ax-b}_2^2
\le
\eps\opt_\cC,
\]
and hence
\[
\norm{A\hat{x}_{\mathrm{proj}}-b}_2^2
\le
(1+\eps)\opt_\cC.
\]
\end{proof}

The next statement gives the corresponding guarantee for our algorithm based on ellipsoid-regularized sampling.

\convexellipsoidregularizedthm*

\begin{proof}[Proof of \Cref{thm:convex-ellipsoid-regularized}]
Let
\[
x^\star \in \argmin_{u\in\cC}\norm{Au-b}_2^2,
\qquad
r\coloneqq Ax^\star-b,
\qquad
x\coloneqq \hat{x}-x^\star.
\]
Write \(\lambda=\lambda_\star\), \(H=A^\top A+\lambda Q\), and
\(d_\lambda=\dlambda(A;Q)\). By the choice of \(Q\) in
\eqref{eq:constraint-induced-dimension}, we have \(d_\lambda\le2\deff(\cC)\).

Let \(c'>0\) be a sufficiently small absolute constant, and set
\(\eta=c'\eps\).
Apply \Cref{lem:one-sided-ridge-subspace-embedding} with constant accuracy
\(1/2\) and failure probability \(\delta/2\). Also apply
\Cref{lem:ridge-amm-residual-cross-term} with accuracy \(\sqrt{\eta}\),
failure probability \(\delta/2\), and fixed vector \(r\). The stated lower
bound on \(m\), with a sufficiently large absolute constant absorbing the
factor \(1/c'\), guarantees that with probability at least \(1-\delta\), both
events hold:
\begin{align}
\label{eq:convex-ellipsoid-one-sided-event}
\norm{SAu}_2^2+\lambda\qnorm{u}{Q}^2
&\ge
\frac12\paren{\norm{Au}_2^2+\lambda\qnorm{u}{Q}^2}
\qquad\text{for all }u\in\RR^d,\\
\label{eq:convex-ellipsoid-amm-event}
\abs{u^\top A^\top(S^\top S-I)r}
&\le
\sqrt{\eta}\sqrt{u^\top Hu}\,\norm{r}_2
\qquad\text{for all }u\in\RR^d.
\end{align}
We condition on these two events for the rest of the proof.

The optimality of \(\hat{x}\) for the sampled constrained problem gives
\[
\norm{S(A\hat{x}-b)}_2^2
\le
\norm{S(Ax^\star-b)}_2^2.
\]
Since \(A\hat{x}-b=Ax+r\), this is equivalent to
\begin{equation}
\label{eq:convex-ellipsoid-sampled-optimality}
\norm{SAx}_2^2+2x^\top A^\top S^\top S r\le0.
\end{equation}
The true excess loss is
\[
G
\coloneqq
\cL(\hat{x})-\opt_\cC
=
\norm{Ax}_2^2+2x^\top A^\top r.
\]
Adding and subtracting the sampled quadratic and linear terms yields
\[
\begin{aligned}
G
&=
\paren{\norm{Ax}_2^2-\norm{SAx}_2^2}
+
2x^\top A^\top(I-S^\top S)r \\
&\qquad
+
\paren{\norm{SAx}_2^2+2x^\top A^\top S^\top S r}.
\end{aligned}
\]
By \eqref{eq:convex-ellipsoid-sampled-optimality}, the last term is
nonpositive. Hence
\begin{equation}
\label{eq:convex-ellipsoid-excess-basic}
G
\le
\norm{Ax}_2^2-\norm{SAx}_2^2
+
2\abs{x^\top A^\top(S^\top S-I)r}.
\end{equation}

Set
\[
X\coloneqq \sqrt{x^\top Hx}
=
\sqrt{\norm{Ax}_2^2+\lambda\qnorm{x}{Q}^2}.
\]
The one-sided embedding event gives
\[
\norm{Ax}_2^2-\norm{SAx}_2^2
\le
\frac12X^2,
\]
and the AMM event gives
\[
2\abs{x^\top A^\top(S^\top S-I)r}
\le
2\sqrt{\eta}\,X\norm{r}_2
=
2\sqrt{\eta}\,X\sqrt{\opt_\cC}
\le
\frac14X^2+4\eta\opt_\cC.
\]
Substituting these bounds into
\eqref{eq:convex-ellipsoid-excess-basic} gives
\begin{equation}
\label{eq:convex-ellipsoid-excess-by-X}
G
\le
\frac34X^2+4\eta\opt_\cC.
\end{equation}

It remains to control \(X\). Since \(\cC\) is convex and \(\cL\) is
differentiable, the first-order optimality condition at \(x^\star\) gives
\[
(z-x^\star)^\top A^\top r\ge0
\qquad\text{for all }z\in\cC.
\]
Taking \(z=\hat{x}\) gives \(x^\top A^\top r\ge0\). Combining this sign
condition with \eqref{eq:convex-ellipsoid-sampled-optimality}, we obtain
\[
\norm{SAx}_2^2
\le
-2x^\top A^\top S^\top S r
=
-2x^\top A^\top r
-2x^\top A^\top(S^\top S-I)r
\le
2\abs{x^\top A^\top(S^\top S-I)r}.
\]
Because \(Q\) is feasible for \eqref{eq:constraint-induced-dimension}, we have
\(\cC\subseteq\cE(Q)\). Thus \(x^\star,\hat{x}\in\cC\) imply
\(\qnorm{x}{Q}\le2\). Using
\eqref{eq:convex-ellipsoid-one-sided-event} and
\eqref{eq:convex-ellipsoid-amm-event}, we get
\[
\frac12X^2
\le
\norm{SAx}_2^2+\lambda\qnorm{x}{Q}^2
\le
2\sqrt{\eta}\,X\sqrt{\opt_\cC}+4\lambda
\le
\frac14X^2+4\eta\opt_\cC+4\lambda.
\]
Therefore
\begin{equation}
\label{eq:convex-ellipsoid-H-bound}
X^2
\le
16\eta\opt_\cC+16\lambda.
\end{equation}
Combining \eqref{eq:convex-ellipsoid-excess-by-X} and
\eqref{eq:convex-ellipsoid-H-bound} yields
\[
G\le16\eta\opt_\cC+12\lambda.
\]
Finally, by the choice
\(\lambda=\lambda_\star\le c_\lambda\frac{\eps}{1+\eps}\opt_\cC\), the
preceding display gives
\[
G\le(16c'+12c_\lambda)\eps\opt_\cC.
\]
Choosing \(c'\) and \(c_\lambda\) sufficiently small gives
\(G\le\eps\opt_\cC\). Hence
\[
\cL(\hat{x})
=
\opt_\cC+G
\le
(1+\eps)\opt_\cC,
\]
as claimed.
\end{proof}

\section{Proofs for lower bounds}
\label{app:lower-bound-proofs}

This appendix contains the full proofs of the lower bounds from
\Cref{sec:lower_bounds}. We treat the nonconvex construction first, and then the
ellipsoid-constrained effective-dimension lower bound.

\subsection{Proof of \Cref{thm:lb-d-eps-assouad}}
\label{app:proof-lower-bound-nonconvex}

\nonconvexlowerboundthm*

\begin{proof}
Let
\[
A
=
\frac{1}{\sqrt{n}}
\begin{bmatrix}
\Id_d\\
\Id_d\\
\vdots\\
\Id_d
\end{bmatrix}
\in \RR^{nd \times d},
\]
where the block \(\Id_d\) is repeated \(n\) times. We index coordinates of \(b\) by
pairs \((t,i)\in[n]\times[d]\). For each \(v \in \set{-1,1}^d\), let \(\mathcal{D}_v\)
be the distribution obtained by drawing independent \(\set{-1,1}\)-valued variables
\(\xi_{t,i}\) satisfying \(\PP(\xi_{t,i}=1)=\frac{1}{2}+8\eps v_i\) and
\(\PP(\xi_{t,i}=-1)=\frac{1}{2}-8\eps v_i\), and then setting
\(b_{t,i} \coloneqq \xi_{t,i}/\sqrt{n}\).
Finally, let \(V\) be uniform on \(\set{-1,1}^d\), and define the mixture
\(\mathcal{D}\) by first drawing \(V\) and then drawing \(b \sim \mathcal{D}_V\).
We will prove a distributional lower bound over deterministic response vectors \(b\), and
then conclude the stated worst-case lower bound by Yao's minimax principle. Accordingly,
it suffices to lower bound deterministic active algorithms against the random choice
\(b \sim \mathcal{D}\).

For any \(x \in \cC=\set{-1,1}^d\), we have
\(\cL(x)=\norm{Ax-b}_2^2=\sum_{i=1}^d \frac{1}{n}\sum_{t=1}^n (x_i-\xi_{t,i})^2\).
Writing \(S_i \coloneqq \sum_{t=1}^n \xi_{t,i}\), the coordinatewise losses are either
\[
\frac{1}{n}\sum_{t=1}^n (-1-\xi_{t,i})^2 = 2+\frac{2S_i}{n}
\qquad \text{or} \qquad
\frac{1}{n}\sum_{t=1}^n (1-\xi_{t,i})^2 = 2-\frac{2S_i}{n}.
\]
Hence the optimal coordinate is \(-1\) when \(S_i/n<0\) and \(1\) when \(S_i/n>0\).

Define the event
\[
\mathcal{E}_i
\coloneqq
\set{
\left|\frac{S_i}{n} - 16\eps V_i\right|
\le 4\eps
}
\qquad
\mathcal{E} \coloneqq \bigcap_{i=1}^d \mathcal{E}_i.
\]
By a standard Chernoff bound, the choice of \(n\) ensures
\(\PP(\mathcal{E}_i^c) \le \frac{1}{100d}\) for each \(i \in [d]\), and thus
\(\PP(\mathcal{E}) \ge 0.99\) by a union bound.

On \(\mathcal{E}\), each coordinate is separated from the decision threshold by at
least \(12\eps\): if \(V_i=-1\), then \(S_i/n \le -12\eps\), while if \(V_i=1\),
then \(S_i/n \ge 12\eps\). Therefore, on \(\mathcal{E}\), the unique coordinatewise
optimizer is \(x_i=V_i\). Let
\(\opt_i \coloneqq \min_{x_i \in \set{-1,1}} \frac{1}{n}\sum_{t=1}^n (x_i-\xi_{t,i})^2\),
so that \(\opt_\cC = \sum_{i=1}^d \opt_i\).
Still on \(\mathcal{E}\), we have \(\opt_i \in [2-40\eps,\,2]\) for every \(i\),
and choosing the wrong value in coordinate \(i\) multiplies the coordinatewise loss by
at least \(\frac{2+24\eps}{2-24\eps} \ge 1+16\eps\).
It follows that if \(\hat{x} \in \set{-1,1}^d\) differs from \(V\) on a set
\(W \subseteq [d]\), then on \(\mathcal{E}\),
\[
\cL(\hat{x})
\ge
\opt_\cC + 16\eps \sum_{i \in W}\opt_i.
\]
Since \(\opt_i \ge (1-20\eps)\opt_\cC/d\) for \(\eps \le 1/400\), this yields
\(\cL(\hat{x}) \ge \paren{1+8\eps\frac{|W|}{d}}\opt_\cC\).
In particular,
\begin{equation}
\label{eq:lb-large-hamming-implies-failure}
\mathcal{E} \cap \set{|W| \ge d/4}
\subseteq
\set{\cL(\hat{x}) \ge (1+2\eps)\opt_\cC}.
\end{equation}

Now fix a deterministic active algorithm, and let \(P_v\) denote its transcript
distribution when \(b \sim \mathcal{D}_v\). This transcript consists of the queried
indices and the observed responses. For \(i \in [d]\), let \(V^{(i)}\)
denote \(V\) with the sign of coordinate \(i\) flipped. Applying
\Cref{lem:weighted-assouad-coordinate-tv} with all weights set to \(1\) gives
\[
\EE\bracket{\sum_{i=1}^d \mathbf{1}\set{\hat{x}_i \neq V_i}}
\ge
\frac{1}{2}\sum_{i=1}^d
\paren{1 - \EE\bracket{D_{\mathrm{TV}}(P_V,P_{V^{(i)}})}}.
\]

To bound the total variation terms, let \(M_i\) be the number of queries made to
coordinates in block \(i\), so that \(\sum_{i=1}^d M_i \le m\) almost surely. Flipping
the sign of coordinate \(i\) in \(V\) changes only the distribution of answers in block
\(i\), and each such answer is \(\set{-1,1}\)-valued with
\(\PP(\xi_{t,i}=1)=1/2-8\eps\) or \(1/2+8\eps\), respectively. By the chain rule for KL divergence for adaptive transcripts, see for
example Lemma 15.1 of \citet{lattimore2020bandit}, this implies
\(D_{\mathrm{KL}}(P_v \,\Vert\, P_{v^{(i)}}) \le C_1 \eps^2 \EE_v[M_i]\)
for an absolute constant \(C_1\). By Pinsker's inequality,
\(D_{\mathrm{TV}}(P_v,P_{v^{(i)}}) \le \sqrt{C_2 \eps^2 \EE_v[M_i]}\)
for another absolute constant \(C_2\). Averaging over \(v\) and using Jensen's
inequality,
\[
\frac{1}{d}\sum_{i=1}^d \EE\bracket{D_{\mathrm{TV}}(P_V,P_{V^{(i)}})}
\le
\sqrt{C_2\eps^2 \frac{1}{d}\sum_{i=1}^d \EE[M_i]}
\le
\sqrt{C_2\eps^2 \frac{m}{d}}.
\]
If \(m \le c\,d/\eps^2\) for \(c>0\) sufficiently small, then the right-hand side is at
most \(1/4\), and therefore
\(\EE\bracket{\sum_{i=1}^d \mathbf{1}\set{\hat{x}_i \neq V_i}}
\ge \frac{d}{2}(1-\frac{1}{4}) = \frac{3d}{8}\).
Let $H \coloneqq |W|$, then \(H = \sum_{i=1}^d \mathbf{1}\set{\hat{x}_i \neq V_i}\). Since
\(0 \le H \le d\), the bound \(\EE[H] \ge 3d/8\) implies
\(\PP(|W| \ge d/4)=\PP(H \ge d/4) \ge \frac{\EE[H]-d/4}{d-d/4} = \frac{1}{6}\).
Combining this with \eqref{eq:lb-large-hamming-implies-failure} and
\(\PP(\mathcal{E}^c)\le 0.01\), we obtain
\(\PP_{b\sim\mathcal{D}}(\cL(\hat{x}) \le (1+2\eps)\opt_\cC)
\le 1-\frac{1}{6}+0.01 < 0.95\).
This proves that every deterministic active algorithm making fewer than \(c\,d/\eps^2\)
queries fails with probability at least \(0.05\) under the above distribution over fixed
vectors \(b\). Yao's minimax principle now implies that every randomized active algorithm
making fewer than \(c\,d/\eps^2\) queries must fail on some fixed response vector
\(b \in \RR^{nd}\), completing the proof.
\end{proof}

\subsection{Proofs of \Cref{thm:effective-dimension-lower-bound} and \Cref{cor:effective-dimension-ball}}
\label{app:proof-lower-bound-effective-dimension}

We first prove the ridge lower bound, where the hard instance is built from a repeated
identity matrix and a sign family supported on the active leverage directions.
We choose coordinate-dependent noise variances according to the regularized leverage
eigenvalues, control the empirical objective for a finite repetition count, and use a
weighted Assouad argument to handle adaptive queries. We then
deduce the ellipsoid-constrained corollary by a Lagrangian reduction.
The ellipsoid radius is fixed before the response vector is drawn.

\effectivedimensionlowerboundthm*

\begin{proof}
Define
\[
A=\frac1{\sqrt n}
\begin{bmatrix}
\Id_d\\
\Id_d\\
\vdots\\
\Id_d
\end{bmatrix}\in\RR^{nd\times d}.
\]
Then \(A^\top A=\Id_d\). Recall that the \((\lambda,Q)\)-ellipsoid-regularized
leverage score of a row \(a_j^\top\) is
\[
\lev_j(A;\lambda,Q)=a_j^\top(A^\top A+\lambda Q)^{-1}a_j.
\]
For the row indexed by \((t,i)\), we have \(a_{t,i}=e_i/\sqrt n\), and hence
\[
\lev_{t,i}(A;\lambda,Q)
=\frac1n e_i^\top(\Id_d+\lambda Q)^{-1}e_i
=\frac1n(1+\lambda q_i)^{-1}.
\]
Thus the per-row score is normalized by \(n\). We write
\(\tau_i\coloneqq(1+\lambda q_i)^{-1}\) for the total ellipsoid leverage mass of the
\(n\) rows in coordinate \(i\), since \(\sum_{t=1}^n\lev_{t,i}(A;\lambda,Q)=\tau_i\).
Equivalently, with \(H=A^\top A+\lambda Q\), we have
\(H^{-1/2}A^\top A H^{-1/2}=(\Id_d+\lambda Q)^{-1}\), so
\(\tau_1,\ldots,\tau_d\) are the eigenvalues in
\Cref{def:truncated-effective-dimension}. Let
\(\mathcal{J}_\eps\coloneqq\Set{i\in[d]}{\tau_i\ge\eps}\), and write
\[
d_\eps\coloneqq\dlambda^\eps(A;Q)
=\sum_{i\in\mathcal{J}_\eps}\tau_i.
\]
If \(d_\eps=0\), the claim is vacuous, so assume \(d_\eps>0\). Let
\(s=|\mathcal{J}_\eps|\) and \(\mu=s^{-1/2}\).

Let \(V\) be uniform on \(\set{-1,1}^{\mathcal{J}_\eps}\). For each
\(i\in\mathcal{J}_\eps\), set \(\sigma_i^2=\mu^2\tau_i/\eps\), draw independent
\(\eta_{t,i}\sim N(0,\sigma_i^2)\), and define
\[
b_{t,i}=\frac{\mu V_i+\eta_{t,i}}{\sqrt n},
\qquad (t,i)\in[n]\times\mathcal{J}_\eps.
\]
For \(i\notin\mathcal{J}_\eps\), set \(b_{t,i}=0\). Once \(V\) and the noises are
drawn, \(b\) is fixed in advance.

For \(i\in\mathcal{J}_\eps\), let \(\overline{\eta}_i=\frac{\sum_t\eta_{t,i}}{n}\) and
\(q_i^\eta=\frac{\sum_t\eta_{t,i}^2}{n}\). Define
\[
\mathcal{E}_i
\coloneqq
\set{\abs{\overline{\eta}_i}\le\mu/4}
\cap
\set{\sigma_i^2/2\le q_i^\eta\le 3\sigma_i^2/2},
\qquad
\mathcal{E}\coloneqq\bigcap_{i\in\mathcal{J}_\eps}\mathcal{E}_i.
\]
Since \(\overline{\eta}_i\sim N(0,\sigma_i^2/n)\) and
\(\sigma_i^2/\mu^2=\tau_i/\eps\le1/\eps\), the elementary Gaussian tail bound
\(\PP(|G|>u)\le2e^{-u^2/2}\) for \(G\sim N(0,1)\) gives
\[
\PP(\abs{\overline{\eta}_i}>\mu/4)
\le2\exp\paren{-\frac{n\mu^2}{32\sigma_i^2}}
\le2\exp(-n\eps/32).
\]
Also, \(q_i^\eta/\sigma_i^2\) is the average of \(n\) independent \(\chi^2_1\) random
variables, so the chi-square concentration bound from Lemma~1 and its specialization in
equations~(4.3)--(4.4) of \citet{LaurentMassart2000} gives
\(\PP(q_i^\eta\notin[\sigma_i^2/2,3\sigma_i^2/2])\le2e^{-c'n}\). Thus, by the choice of
\(n\) and a union bound, \(\PP(\mathcal{E})\ge0.99\).

Fix a realization in \(\mathcal{E}\). For \(x\in\RR^d\), the objective decomposes as
the sum of the active-coordinate terms
\(n^{-1}\sum_t(x_i-\mu V_i-\eta_{t,i})^2+\lambda q_i x_i^2\) for
\(i\in\mathcal{J}_\eps\), and the inactive-coordinate terms
\((1+\lambda q_i)x_i^2\). Completing the square in an active coordinate gives
\begin{equation}
\label{eq:anisotropic-coordinate-square-completion}
\frac1n\sum_{t=1}^n(x_i-\mu V_i-\eta_{t,i})^2+\lambda q_i x_i^2
=(1+\lambda q_i)(x_i-x_i^\star)^2+r_i,
\end{equation}
where
\[
x_i^\star=\tau_i(\mu V_i+\overline{\eta}_i),
\qquad
r_i=q_i^\eta-\overline{\eta}_i^2+(1-\tau_i)(\mu V_i+\overline{\eta}_i)^2.
\]
For \(i\notin\mathcal{J}_\eps\), the coordinate minimizer is \(x_i^\star=0\). Hence
\(x^\star\) is the ridge minimizer and
\[
\opt_{\lambda,Q}(A,b)=\sum_{i\in\mathcal{J}_\eps}r_i.
\]
On \(\mathcal{E}\), for every \(i\in\mathcal{J}_\eps\),
\[
\abs{x_i^\star}
=\tau_i\abs{\mu V_i+\overline{\eta}_i}
\ge \tau_i\paren{\mu-\abs{\overline{\eta}_i}}
\ge \frac34\mu\tau_i,
\]
and therefore \(\mathrm{sign}(x_i^\star)=V_i\). Moreover,
\[
r_i\le \frac32\sigma_i^2+\paren{\frac54\mu}^2
\le 4\,\frac{\mu^2\tau_i}{\eps},
\]
where the last step uses \(\tau_i\ge\eps\). Therefore
\(\opt_{\lambda,Q}(A,b)\le4\mu^2d_\eps/\eps\).

For any \(\hat{x}\), define the weighted sign error
\[
H_\tau(\hat{x},V)
\coloneqq
\sum_{i\in\mathcal{J}_\eps}\tau_i\,\mathbf{1}\set{\mathrm{sign}(\hat{x}_i)\neq V_i},
\]
with arbitrary tie-breaking at zero. Let
\(W\coloneqq\Set{i\in\mathcal{J}_\eps}{\mathrm{sign}(\hat{x}_i)\neq V_i}\). For
\(i\in W\), the point \(x_i^\star\) has sign \(V_i\), while \(\hat{x}_i\) is either on
the opposite side of zero or is tied at zero with the wrong decoded sign. Thus
\begin{equation}
\label{eq:anisotropic-sign-mistake-distance}
\abs{\hat{x}_i-x_i^\star}
\ge \abs{x_i^\star}
\ge \frac34\mu\tau_i.
\end{equation}
Using \eqref{eq:anisotropic-coordinate-square-completion} and the display for
\(\opt_{\lambda,Q}(A,b)\), the excess loss satisfies
\begin{align}
\label{eq:anisotropic-ridge-excess-coordinate-decomposition}
\cL_{\lambda,Q}(\hat{x};b)-\opt_{\lambda,Q}(A,b)
\;&=
\sum_{i\in\mathcal{J}_\eps}(1+\lambda q_i)(\hat{x}_i-x_i^\star)^2
\;+\!\sum_{i\notin\mathcal{J}_\eps}(1+\lambda q_i)\hat{x}_i^2 \nonumber\\
&\ge
\sum_{i\in W}(1+\lambda q_i)(\hat{x}_i-x_i^\star)^2.
\end{align}
Since \(1+\lambda q_i=1/\tau_i\), \eqref{eq:anisotropic-sign-mistake-distance} shows
that each \(i\in W\) contributes at least
\((1/\tau_i)(3\mu\tau_i/4)^2=(9\mu^2/16)\tau_i\). Summing over \(W\) yields
\[
\cL_{\lambda,Q}(\hat{x};b)-\opt_{\lambda,Q}(A,b)
\ge
\frac{9\mu^2}{16}H_\tau(\hat{x},V).
\]
Combining this with the upper bound on \(\opt_{\lambda,Q}(A,b)\), and choosing
\(c_0\le 9/256\), yields
\begin{equation}
\label{eq:anisotropic-weighted-error-implies-ridge-failure}
\mathcal{E}\cap\set{H_\tau(\hat{x},V)\ge d_\eps/4}
\subseteq
\set{\cL_{\lambda,Q}(\hat{x};b)\ge(1+c_0\eps)\opt_{\lambda,Q}(A,b)}.
\end{equation}

It remains to show that, under the hard distribution, any algorithm with too few queries
must make many sign mistakes in the weighted sense measured by \(H_\tau\). Fix a
deterministic active algorithm, and let \(P_v\) denote its transcript distribution under
sign vector \(v\). Let \(M_i\) be the number of queries made in block \(i\). Flipping sign
\(i\) changes only answers in that block, and each such Gaussian answer has variance
\(\sigma_i^2/n\) and means separated by \(2\mu/\sqrt n\). Thus the KL divergence per
queried label in block \(i\) is \(2\mu^2/\sigma_i^2=2\eps/\tau_i\). The adaptive KL chain
rule gives
\[
D_{\mathrm{KL}}(P_v\,\Vert\,P_{v^{(i)}})
\le
\frac{2\eps}{\tau_i}\EE_v[M_i].
\]
Pinsker's inequality and Jensen's inequality imply
\[
\EE[D_{\mathrm{TV}}(P_V,P_{V^{(i)}})]
\le
\sqrt{\frac{\eps\,\EE[M_i]}{\tau_i}},
\]
where the expectation averages over \(V\). Identifying the active coordinate set
\(\mathcal{J}_\eps\) with \([s]\), apply
\Cref{lem:weighted-assouad-coordinate-tv} with weights \(w_i=\tau_i\) to get
\begin{align}
\label{eq:anisotropic-assouad-weighted-sign-error}
\EE\bracket{H_\tau(\hat{x},V)}
\;&\ge
\frac12\sum_{i\in\mathcal{J}_\eps}
\tau_i\paren{1-\EE\bracket{D_{\mathrm{TV}}(P_V,P_{V^{(i)}})}} \nonumber\\
&=
\frac12d_\eps
-\frac12\sum_{i\in\mathcal{J}_\eps}
\tau_i\,\EE\bracket{D_{\mathrm{TV}}(P_V,P_{V^{(i)}})} \nonumber\\
&\ge
\frac12d_\eps
-\frac12\sqrt{\eps}
\sum_{i\in\mathcal{J}_\eps}\sqrt{\tau_i\,\EE[M_i]}.
\end{align}
Finally, Cauchy--Schwarz gives
\[
\sum_{i\in\mathcal{J}_\eps}\sqrt{\tau_i\,\EE[M_i]}
\le
\sqrt{\paren{\sum_{i\in\mathcal{J}_\eps}\tau_i}
\paren{\sum_{i\in\mathcal{J}_\eps}\EE[M_i]}}
\le
\sqrt{d_\eps m},
\]
because the algorithm makes at most \(m\) queries total. Substituting this into
\eqref{eq:anisotropic-assouad-weighted-sign-error} yields
\[
\EE\bracket{H_\tau(\hat{x},V)}
\ge
\frac12d_\eps-\frac12\sqrt{\eps d_\eps m}.
\]
If \(m\le cd_\eps/\eps\) for \(c>0\) sufficiently small, then
\(\EE[H_\tau(\hat{x},V)]\ge0.49d_\eps\). Since \(0\le H_\tau(\hat{x},V)\le d_\eps\),
this implies \(\PP(H_\tau(\hat{x},V)\ge d_\eps/4)>0.3\). Combining this with
\(\PP(\mathcal{E})\ge0.99\) and
\eqref{eq:anisotropic-weighted-error-implies-ridge-failure}, the deterministic algorithm
fails with probability greater than \(0.25\) under the hard distribution. Yao's minimax
principle gives the stated worst-case response vector \(b\).
\end{proof}

We now convert the ridge lower bound into a lower bound for constrained least-squares
over an ellipsoid.

\effectiveballlowerboundcor*

\begin{proof}
If \(\dlambda^\eps(A;Q)=0\), take any \(R>0\); the claim is vacuous.
Hence assume \(\dlambda^\eps(A;Q)>0\).
Use the same matrix \(A\), active set \(\mathcal{J}_\eps\), values
\(\tau_i=(1+\lambda q_i)^{-1}\), and hard distribution over \(b\) as in the proof of
\Cref{thm:effective-dimension-lower-bound}. Let \(s=|\mathcal{J}_\eps|\),
\(\mu=s^{-1/2}\), and \(d_\eps=\sum_{i\in\mathcal{J}_\eps}\tau_i\). Fix, before drawing
the response vector, the radius
\[
R\coloneqq \frac{25}{16}\mu^2\sum_{i\in\mathcal{J}_\eps}q_i\tau_i^2,
\qquad
\cC\coloneqq\Set{x\in\RR^d}{x^\top Qx\le R}.
\]
Let \(\mathcal{E}\) be the concentration event from the theorem proof, and set
\(y_i=\mu V_i+\overline{\eta}_i\) on \(\mathcal{J}_\eps\), and \(y_i=0\) otherwise. On
\(\mathcal{E}\), \(\abs{y_i}\le5\mu/4\), so the ridge minimizer
\(x_i^\lambda=\tau_i y_i\) is feasible for \(\cC\). Hence
\(\opt_\cC\le\norm{Ax^\lambda-b}_2^2\le\opt_{\lambda,Q}(A,b)\le4\mu^2d_\eps/\eps\).

The only new step is to replace the ridge sign-error calculation by a constrained one.
Let \(x^\cC=(x_i^\cC)_{i=1}^d\in\argmin_{x\in\cC}\norm{Ax-b}_2^2\). Since
\(A^\top A=\Id_d\) and \(Q\) is diagonal, the KKT conditions give a multiplier
\(\kappa\ge0\) with
\(x_i^\cC=y_i/(1+\kappa q_i)\). Also \(\kappa\le\lambda\), because the squared
\(Q\)-norm of \((y_i/(1+\gamma q_i))_i\) is nonincreasing in \(\gamma\), and the point
corresponding to \(\gamma=\lambda\) is feasible. For any \(\hat{x}\in\cC\), completing
the square at multiplier \(\kappa\) and using feasibility gives
\[
\norm{A\hat{x}-b}_2^2-\opt_\cC
\ge
\sum_{i=1}^d(1+\kappa q_i)(\hat{x}_i-x_i^\cC)^2.
\]
On \(\mathcal{E}\), \(\abs{y_i}\ge3\mu/4\) and \(\mathrm{sign}(y_i)=V_i\). Thus, for
each sign mistake \(i\in\mathcal{J}_\eps\),
\[
(1+\kappa q_i)(\hat{x}_i-x_i^\cC)^2
\ge
(1+\kappa q_i)(x_i^\cC)^2
=
\frac{y_i^2}{1+\kappa q_i}
\ge
\frac{9\mu^2}{16}\tau_i,
\]
where the last step uses \(\kappa\le\lambda\), so
\((1+\kappa q_i)^{-1}\ge(1+\lambda q_i)^{-1}=\tau_i\). Therefore
\[
\norm{A\hat{x}-b}_2^2-\opt_\cC
\ge
\frac{9\mu^2}{16}H_\tau(\hat{x},V).
\]
Combining with the above upper bound on \(\opt_\cC\), on
\(\mathcal{E}\cap\set{H_\tau(\hat{x},V)\ge d_\eps/4}\),
\[
\norm{A\hat{x}-b}_2^2
\ge
\paren{1+\frac{9}{256}\eps}\opt_\cC.
\]

The transcript lower bound from the theorem proof applies unchanged to algorithms for
this fixed ellipsoid, and gives the event
\(\mathcal{E}\cap\set{H_\tau(\hat{x},V)\ge d_\eps/4}\) with probability greater than
\(0.25\) whenever the query budget is below \(c\,\dlambda^\eps(A;Q)/\eps\). Yao's
principle gives a fixed hard response vector. Finally,
\(A^\top A+\lambda R(R^{-1}Q)=A^\top A+\lambda Q\), so
\(d_{\lambda R}^\eps(A;R^{-1}Q)=\dlambda^\eps(A;Q)\), and the result follows after
choosing \(c_0\le9/256\).
\end{proof}

\section{Additional technical lemmas}
\label{app:additional-technical-lemmas}

\begin{lemma}[Assouad for weighted sign error]
\label{lem:weighted-assouad-coordinate-tv}
Let \(\set{P_v}_{v\in\set{-1,1}^m}\) be a family of distributions on an observation
space \(\mathcal{T}\). Let \(V\) be uniform on \(\set{-1,1}^m\), let \(T\sim P_V\),
and let \(\hat{v}=\hat{v}(T)\in\set{-1,1}^m\) be any estimator. For each \(i\in[m]\),
let \(V^{(i)}\) denote \(V\) with coordinate \(i\) flipped. Then for any nonnegative
weights \(w_1,\ldots,w_m\),
\[
\EE\bracket{\sum_{i=1}^m w_i\mathbf{1}\set{\hat{v}_i\neq V_i}}
\ge
\frac12\sum_{i=1}^m
w_i\paren{1-\EE\bracket{D_{\mathrm{TV}}(P_V,P_{V^{(i)}})}}.
\]
\end{lemma}

\begin{proof}
It suffices to prove the coordinatewise bound and sum with weights. Fix \(i\), and
condition on \(V^{-i}=u\). Let \(v^-\) and \(v^+\) be the two vertices extending \(u\)
with \(i\)-th coordinate \(-1\) and \(1\), respectively. Since \(V_i\) is conditionally
uniform,
\[
\PP(\hat{v}_i\neq V_i\mid V^{-i}=u)
=
\frac12\PP_{P_{v^-}}(\hat{v}_i=1)
+
\frac12\PP_{P_{v^+}}(\hat{v}_i=-1).
\]
For \(E=\set{\hat{v}_i=1}\), total variation gives
\(\PP_{P_{v^+}}(E)-\PP_{P_{v^-}}(E)\le D_{\mathrm{TV}}(P_{v^-},P_{v^+})\). Hence
\[
\PP(\hat{v}_i\neq V_i\mid V^{-i}=u)
\ge
\frac12\paren{1-D_{\mathrm{TV}}(P_{v^-},P_{v^+})}.
\]
Averaging over \(u\) is the same as averaging uniformly over all cube edges in direction
\(i\). Since total variation is symmetric, this gives
\[
\PP(\hat{v}_i\neq V_i)
\ge
\frac12\paren{1-\EE\bracket{D_{\mathrm{TV}}(P_V,P_{V^{(i)}})}}.
\]
Multiplying by \(w_i\) and summing over \(i\) proves the result.
\end{proof}

\section{Comparison with prior lower bounds}
\label{app:lower-bound-comparison}

The arXiv version
\citep{DevvritRajaramanAwasthi2021v1} and the published NeurIPS version
\citep{rajaraman2022semi} of the semi-supervised active regression paper contain different
lower bounds. We compare each with \Cref{thm:effective-dimension-lower-bound}.

First, compared with the lower bound of
\citet[Theorem~4]{DevvritRajaramanAwasthi2021v1}, our theorem makes the repetition count
\(n\) explicit, while their proof takes \(n\to\infty\) without specifying
a finite value. We also extend their isotropic case \(Q=\Id_d\), restricted to
\(\lambda\in[1,50]\) and \(\eps\le1/100\), to arbitrary diagonal \(Q\succ0\),
every \(\lambda\ge0\), and \(\eps\in(0,1]\), yielding the anisotropic truncated effective dimension
\(\dlambda^\eps(A;Q)\).

Second, compared with the lower bound of
\citet[Theorem~5]{rajaraman2022semi}, our theorem concerns constant-probability
approximation of the empirical objective on a fixed finite pool. Their result concerns
distributional regression with fresh noise at each observation and expected
generalization loss. Under the regularized specialization
\(X_{\mathrm{un}}=A\), \(X_{\mathrm{lab}}=\sqrt{\lambda}\,Q^{1/2}\), with zero labeled
responses and \(A^\top A=\Id_d\), their additional spectral assumption becomes
\(\min_i\sqrt{\lambda q_i/(1+\lambda q_i)}\ge\eps/(1-\eps)\), excluding
\(\lambda=0\) and sufficiently weakly regularized directions. Our theorem requires
no such condition.

Finally, \citet[Theorem~28]{ChenPrice:2019a} prove an
\(\Omega(d/\eps)\) lower bound for unregularized regression, even with adaptive queries.
They also give a condition-number-dependent lower bound
\citep[Theorem~31]{ChenPrice:2019a}, but under i.i.d.\ sampling rather than actively
chosen queries. Nevertheless, these settings share a common lower-bound technique:
the arXiv ridge proof \citep{DevvritRajaramanAwasthi2021v1} adapts their packing and
information argument \citep[Claim~29 and proof of Theorem~28]{ChenPrice:2019a}.

\providecommand{\ellipsoidsdpsection}{
\begingroup

\section{Computing the enclosing ellipsoid}
\label{app:ellipsoid-sdp}

The choice of an enclosing ellipsoid depends on the constraint set and how it is specified.

For finite feasible sets and polytopes specified by vertices, minimizing the total
ellipsoid-regularized leverage-score mass admits a semidefinite formulation.
Let $\cC=\{x_1,\ldots,x_N\}$ or
$\cC=\operatorname{conv}\{x_1,\ldots,x_N\}$, in centered coordinates.
Since an ellipsoid is convex, containment of $\cC$ is equivalent to requiring
$x_i^\top Qx_i\le1$ for every $i\in[N]$.
Thus, writing $B=A^\top A\succ0$ and fixing $\lambda=\lambda_\star\ge0$,
Equation~\eqref{eq:constraint-induced-dimension} becomes
\[
\min_{Q\succeq0}\ \tr\!\left(B(B+\lambda Q)^{-1}\right)
\quad\text{subject to}\quad
x_i^\top Qx_i\le1,\qquad i\in[N].
\]

This problem has the following semidefinite program formulation:
\begin{equation}
\label{eq:ellipsoid-sdp}
\begin{aligned}
    \min_{Q,Y}\quad & \tr(Y)\\
    \text{subject to}\quad
        & \begin{pmatrix}
            B+\lambda Q & B^{1/2}\\
            B^{1/2} & Y
          \end{pmatrix}\succeq0,\\
        & x_i^\top Qx_i\le1,\qquad i\in[N],\qquad Q\succeq0.
\end{aligned}
\end{equation}
By the Schur complement, the block constraint is equivalent to
\[
Y\succeq B^{1/2}(B+\lambda Q)^{-1}B^{1/2},
\]
with equality at the optimum for each fixed $Q$.
Here, minima are interpreted as infima if not attained; allowing $Q\succeq0$
instead of $Q\succ0$ does not change the infimum in
\eqref{eq:constraint-induced-dimension}.
This applies, for example, to finite feasible sets, simplices, and polytopes
specified by a finite list of vertices.

For more general constraint sets, a minimum-volume enclosing ellipsoid is a possible
heuristic. Computing and approximating such ellipsoids is a well-studied
problem~\citep{CohenCousinsLeeYang2019,WoodruffYasuda2024,GulerGurtuna2012}.

\endgroup
}

\begingroup

\section{Computing the enclosing ellipsoid}
\label{app:ellipsoid-sdp}

The choice of an enclosing ellipsoid depends on the constraint set and how it is specified.

For finite feasible sets and polytopes specified by vertices, minimizing the total
ellipsoid-regularized leverage-score mass admits a semidefinite formulation.
Let $\cC=\{x_1,\ldots,x_N\}$ or
$\cC=\operatorname{conv}\{x_1,\ldots,x_N\}$, in centered coordinates.
Since an ellipsoid is convex, containment of $\cC$ is equivalent to requiring
$x_i^\top Qx_i\le1$ for every $i\in[N]$.
Thus, writing $B=A^\top A\succ0$ and fixing $\lambda=\lambda_\star\ge0$,
Equation~\eqref{eq:constraint-induced-dimension} becomes
\[
\inf_{Q\succeq0}\ \tr\!\left(B(B+\lambda Q)^{-1}\right)
\quad\text{subject to}\quad
x_i^\top Qx_i\le1,\qquad i\in[N].
\]

This problem has the following semidefinite program formulation:
\begin{equation}
\label{eq:ellipsoid-sdp}
\begin{aligned}
    \inf_{Q,Y}\quad & \tr(Y)\\
    \text{subject to}\quad
        & \begin{pmatrix}
            B+\lambda Q & B^{1/2}\\
            B^{1/2} & Y
          \end{pmatrix}\succeq0,\\
        & x_i^\top Qx_i\le1,\qquad i\in[N],\qquad Q\succeq0.
\end{aligned}
\end{equation}
By the Schur complement, the block constraint is equivalent to
\[
Y\succeq B^{1/2}(B+\lambda Q)^{-1}B^{1/2},
\]
with equality at the optimum for each fixed $Q$.
This applies, for example, to finite feasible sets, simplices, and polytopes
specified by a finite list of vertices.

For more general constraint sets, a minimum-volume enclosing ellipsoid is a possible
heuristic. Computing and approximating such ellipsoids is a well-studied
problem~\citep{CohenCousinsLeeYang2019,WoodruffYasuda2024,GulerGurtuna2012}.

\endgroup

\section{Additional experiments}
\label{app:additional-experiments}

In this section, we study how constraint anisotropy, the choice of outer
ellipsoid, and the choice of $\lambda$ affect the sampling method. We
first give the common protocol, then describe these comparisons and the
construction details for the main experiments.

\subsection{Sampling and evaluation}
\label{app:experiment-protocol}

Each comparison uses the same feasible set $\cC$ and varies the row
sampling probabilities: uniform, or proportional to $\ell_i(A)$,
$\ell_i^\lambda(A)$, or $\lev_i(A;\lambda,Q)$. We sample with replacement
and compute $\hat x\in\argmin_{x\in\cC}\|S(Ax-b)\|_2^2$, using the
sampling-and-rescaling matrix $S$ from \eqref{eq:sampling-matrix}.
Repeated draws reuse labels and count toward the sample size $m$.
Unless $\lambda$ is varied or estimated from queried labels, we set
$\lambda=\eps\opt_\cC/(1+\eps)$ with $\eps=0.1$ to compare sampling
distributions at the same scale. We compute the constrained fits and
$\opt_\cC$ using CVXPY and Clarabel, and evaluate $\cL(\hat x)$ on all rows.

We fix the design and signal and draw fresh independent label noise
and sampling indices in each trial; the diagonal experiment also reshuffles
rows. All methods and sample sizes share the response vector within a
trial. For the PDE experiment, the design and noisy labels are fixed and
only sampling indices vary. Every plot reports medians and interquartile
ranges across trials.

\subsection{Isolating the effect of constraint anisotropy}
\label{app:experiment-diagonal}

To isolate how the constraint changes label selection, we use $n=5000$
and $d=100$ and give every coordinate equal representation in the design.
We repeat each row $\sqrt{d/n}\,e_j^\top$ 50 times and randomly permute
the rows, so $A^\top A=\Id_d$. We choose
$\cC=\cE(Q)$ with $Q=\diag(10\Id_{95},0.1\Id_5)$, allowing greater
variation in the last five coordinates. The signal
$\theta^\star=\sqrt2\sum_{j=96}^{100}e_j$ lies on the ellipsoid boundary
in these coordinates. We set $b=A\theta^\star+\xi$, with independent
$\xi_i\sim\mathcal N(0,0.1^2)$. We compare the methods over 30 trials
at sample sizes $m\in\{100,200,300,400,500\}$.

\begin{figure}[!htbp]
    \centering
    \begin{minipage}[c][0.31\linewidth][c]{0.76\linewidth}
    \begin{subfigure}[t]{0.48\linewidth}
        \centering
        \includegraphics[width=\linewidth]{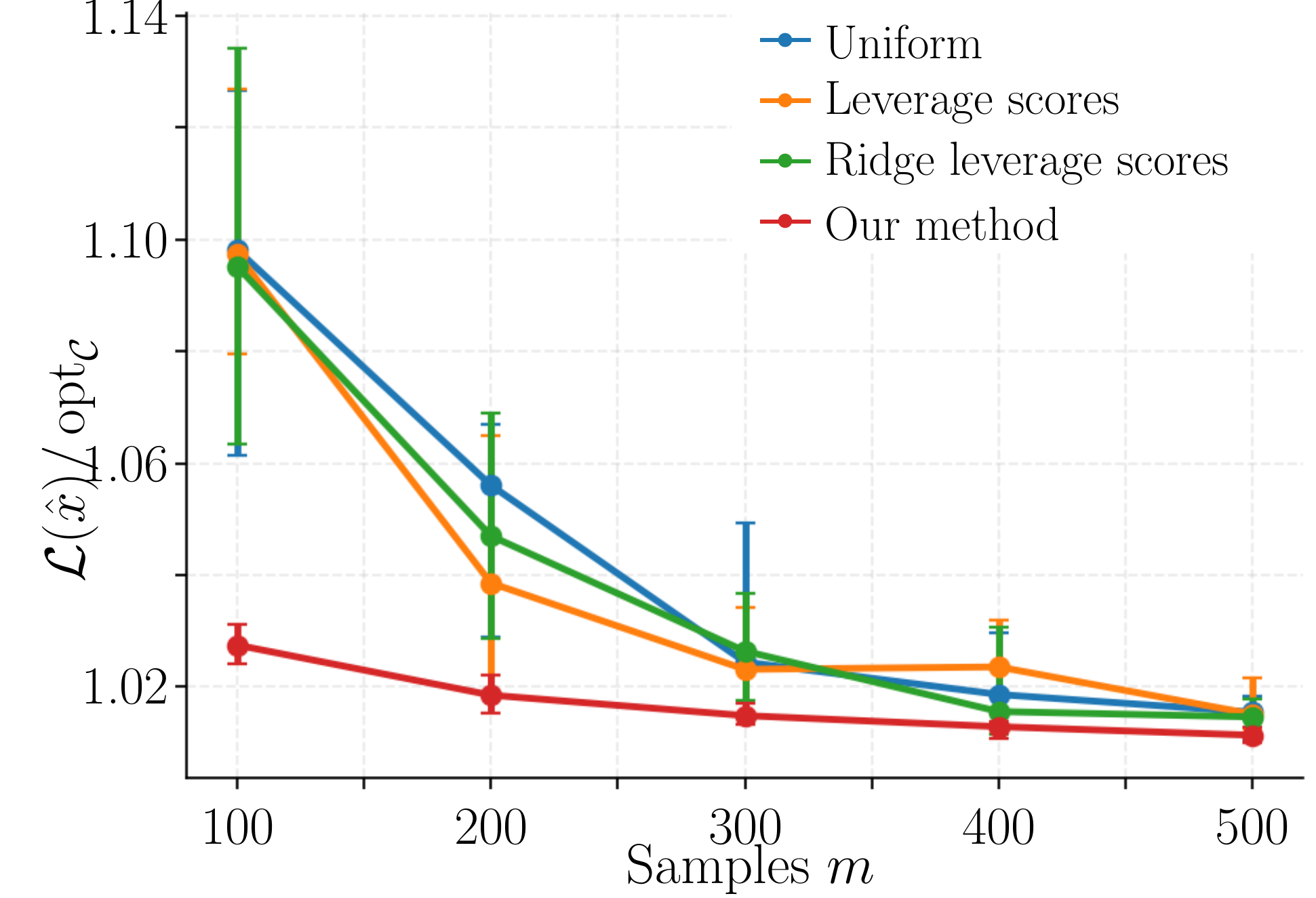}
        \caption{Sampling distributions.}
        \label{fig:coordinate-baselines}
    \end{subfigure}\hfill
    \begin{subfigure}[t]{0.48\linewidth}
        \centering
        \includegraphics[width=\linewidth]{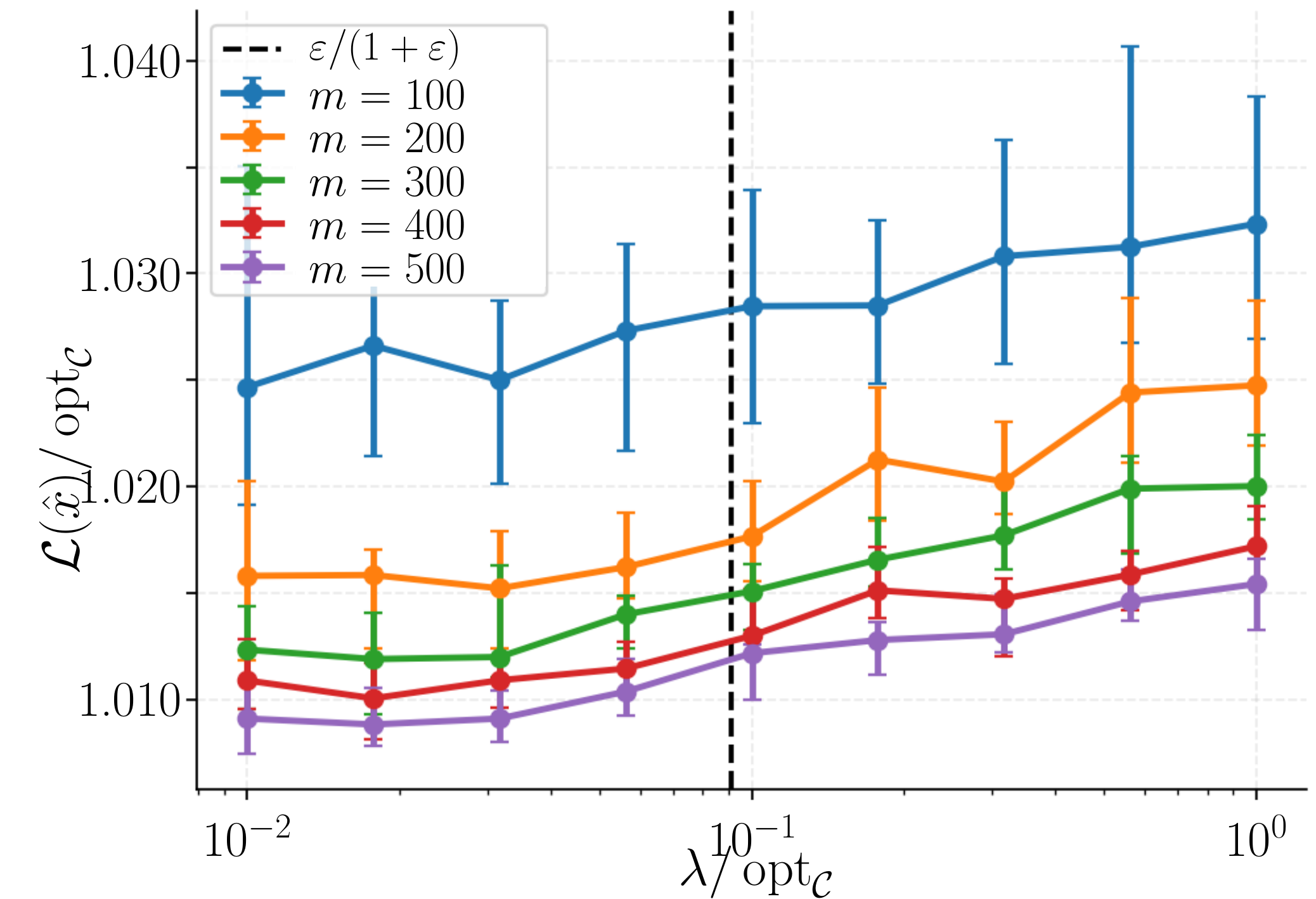}
        \caption{Sensitivity to $\lambda$.}
        \label{fig:coordinate-lambda}
    \end{subfigure}
    \end{minipage}
    \caption{Diagonal constrained regression over 30 trials.
    (a) $\cL(\hat x)/\opt_\cC$ for each sampling distribution.
    (b) Our method as $\lambda/\opt_\cC$ varies; the dashed line marks
    $\eps/(1+\eps)$.}
    \label{fig:coordinate-experiments}
\end{figure}

For a row observing coordinate $j$, the score is
$\lev_i(A;\lambda,Q)=(d/n)/(1+\lambda Q_{jj})$.
Thus the constraint directs samples toward the coordinates carrying the
signal. This gives a lower median of $\cL(\hat x)/\opt_\cC$ in
\Cref{fig:coordinate-experiments}(a), with the greatest advantage at
small sample sizes. To test stability to misspecifying $\lambda$, we also
sweep $\lambda/\opt_\cC=10^{-2+j/4}$, $j=0,\ldots,8$, at each sample
size. The median stays below $1.04$ throughout this range
(panel (b)), suggesting that accurate approximation does not require
precise tuning in this setting.

\subsection{Sensitivity to loose or misaligned outer ellipsoids}
\label{app:experiment-robustness}

To study how accurate the geometric information must be, we use the
ellipsoidal constraint experiment from \Cref{sec:experiments}, with
$n=3000$, $d=60$, and $m=90$. We keep $\cC=\cE(Q)$ and its 50 trial response vectors fixed,
but sample using progressively looser outer ellipsoids with shape matrices
\[
    \widetilde Q_t=(1-t)Q+tQ_{\mathrm{end}},
    \qquad t\in\{0,0.25,0.5,0.75,1\}.
\]
We use two endpoints. The first is
$Q_{\mathrm{end}}=\lambda_{\min}(Q)\Id$, corresponding to the smallest
centered Euclidean ball containing $\cC$. For the second, we draw an
independent random orthogonal matrix $W$ as in
\Cref{app:experiment-geometry-details} and set
\[
    M=W\diag(0.1\Id_6,10\Id_{54})W^\top,
    \qquad Q_{\mathrm{end}}=
    \frac{M}{\lambda_{\max}(Q^{-1/2}MQ^{-1/2})}.
\]
This changes the orientation while ensuring $Q_{\mathrm{end}}\preceq Q$.
Both choices give $\widetilde Q_t\preceq Q$, so every sampling ellipsoid
contains $\cC$. We draw new sampling indices for each $t$ and fit over
the original feasible set.

\begin{figure}[!htbp]
    \centering
    \includegraphics[width=0.76\linewidth]{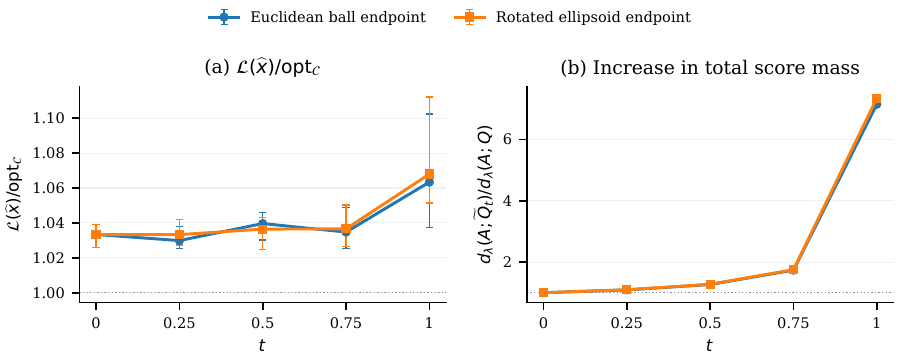}
    \caption{Varying the outer ellipsoid over 50 trials with $m=90$.
    Curves interpolate from $Q$ to the Euclidean ball and rotated ellipsoid
    endpoints. (a) $\cL(\hat x)/\opt_\cC$.
    (b) $d_\lambda(A;\widetilde Q_t)/d_\lambda(A;Q)$.}
    \label{fig:experiment-robustness}
\end{figure}

In \Cref{fig:experiment-robustness}(a), the median of
$\cL(\hat x)/\opt_\cC$ changes little through $t=0.75$ and increases
at $t=1$, remaining below $1.07$ for both endpoints. The total score mass
increases as the ellipsoid becomes larger (panel (b)), consistent with
the monotonicity discussed after \eqref{eq:regularized-score-mass}.
Thus the approximation remains accurate at this sample size even when
the sampler receives a substantially looser description of the constraint.

\subsection{Choosing \texorpdfstring{$\lambda$}{lambda} from queried labels}
\label{app:experiment-lambda}

To test whether a small preliminary sample can provide a useful choice
of $\lambda$, we use $n=2000$, $d=40$, and five long axes in the
construction of \Cref{app:experiment-geometry-details}. We compare methods
over 50 trials at total sample sizes $m\in\{200,250,300,400,500\}$.
\begin{figure}[!htbp]
    \centering
    \includegraphics[width=0.76\linewidth]{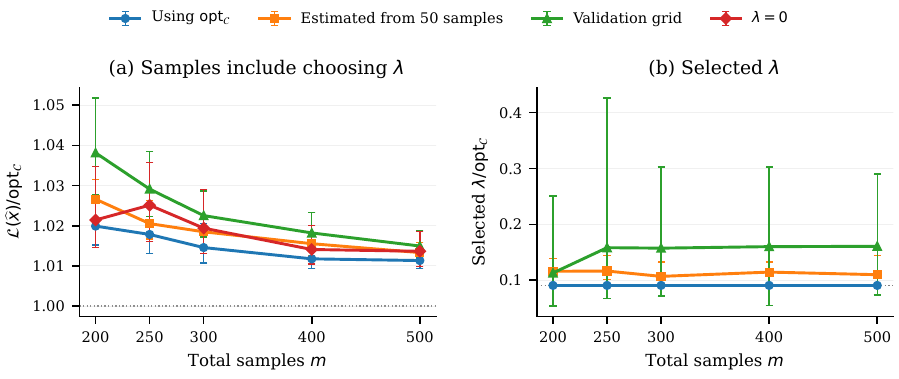}
    \caption{Choosing $\lambda$ from queried labels over 50 trials.
    (a) $\cL(\hat x)/\opt_\cC$ at matched total sample sizes.
    (b) $\lambda/\opt_\cC$; the dotted line marks $\eps/(1+\eps)$.
    The comparison using $\opt_\cC$ and evaluation of $\cL(\hat x)$ use all labels.}
    \label{fig:experiment-lambda}
\end{figure}

We first fit a feasible $\widetilde x$ from 20 rows sampled by leverage
scores. We then draw 30 independent indices $J_t$ with probabilities
$p_i=\ell_i(A)/d$ and set
\[
    \lambda=\frac{\eps}{1+\eps}
    \left(\frac1{30}\sum_{t=1}^{30}
    \frac{(a_{J_t}^\top\widetilde x-b_{J_t})^2}{p_{J_t}}\right).
\]
The average estimates $\cL(\widetilde x)$ without bias conditional on
$\widetilde x$. We use the remaining $m-50$ draws to sample by
$\lev_i(A;\lambda,Q)$ and compute the final fit. We compare with
$\lambda=\eps\opt_\cC/(1+\eps)$ and $\lambda=0$, both of which use
all $m$ draws for fitting.

To compare this estimate with more extensive tuning, we also test a
validation grid with multipliers $\{0.25,0.5,1,2,4\}$ of the estimated
$\lambda$. After the same 50-draw pilot, each candidate uses eight fitting
draws; 40 fresh leverage-score draws select the candidate with the
smallest reweighted validation loss. The final fit uses $m-130$ further
draws at the selected $\lambda$ and reuses that candidate's eight fitting
draws. Thus every method is charged for all draws used in selection.

In \Cref{fig:experiment-lambda}, the 50-sample estimate gives a median
of $\cL(\hat x)/\opt_\cC$ close to that obtained using $\opt_\cC$,
with the gap generally shrinking as $m$ grows. It exchanges ranks with
$\lambda=0$, while the validation grid has a higher median throughout.
At these sample sizes, the simpler estimate leaves more labels for the
final fit and gives a better approximation than the validation grid.

\subsection{Construction details for the main experiments}
\label{app:experiment-constructions}

\subsubsection{Ellipsoidal constraints}
\label{app:experiment-geometry-details}
For the main ellipsoidal experiment and the experiment choosing $\lambda$,
we use $(n,d,k)=(3000,60,6)$ and $(2000,40,5)$, respectively, where $k$
is the number of long axes. The main experiment uses 50 trials at
$m\in\{60,90,120,180,240,360\}$. We obtain a random orthogonal $U$ by QR
factorization of a matrix of independent standard Gaussians, choosing
column signs so that the triangular factor has positive diagonal.
We set $s_j=0.85$ for $j\le k$, $s_j=1.15$ for $k<j\le2k$, and
$s_{2k+r}=0.8+0.4(r-1)/(d-2k-1)$ for $r\in[d-2k]$.
Repeating each row $\sqrt{ds_j/n}\,(Ue_j)^\top$ exactly $n/d$ times
and randomly permuting gives $A^\top A=U\diag(s_1,\ldots,s_d)U^\top$.

We let $R$ rotate each ordered pair $(e_j,e_{j+k})$, $j\in[k]$, by
$35^\circ$ and act as the identity elsewhere. We then set
\[
    Q=UR\diag(0.1\Id_k,10\Id_{d-k})R^\top U^\top,
    \qquad
    \theta^\star=\frac{\sqrt{10}[UR e_1,\ldots,UR e_k]g}{\|g\|_2},
    \qquad g\sim\mathcal N(0,\Id_k).
\]
This places the signal uniformly on the boundary in the span of the long
axes. Labels are $b=A\theta^\star+\xi$, with independent
$\xi_i\sim\mathcal N(0,0.1^2)$.

\subsubsection{PDE, parameter pool, and polynomial features}
\label{app:experiment-pde-details}
For the heat-conduction problem in \Cref{sec:experiments}, we use
$n=2000$ and $d=35$ and compare methods over 120 trials at
$m\in\{30,35,40,50,60,75,100,150,250\}$. The conductivity is
\[
    \kappa(s;\mu)=\exp\!\left(
    0.35\mu_1\sin(\pi s)+0.25\mu_2\cos(2\pi s)+0.15\mu_3\sin(3\pi s)\right).
\]
Each parameter controls a spatial variation in log conductivity.
We solve the PDE on 80 interior points with spacing $h=1/81$, using
centered flux differences with conductivity evaluated at cell midpoints
and boundary values $u_0=u_{81}=0$. The numerical midpoint response is
$q(\mu)=(u_{40}+u_{41})/2$.
The pool contains 1800 samples from $\mathcal N(0,0.22^2\Id_3)$ clipped
coordinatewise to $[-1,1]^3$ and 200 uniform samples from that cube,
randomly permuted. This combines concentration near the origin with
coverage of more distant parameter values.

For nonnegative integer triples $\alpha$ with $|\alpha|\le4$, the 35
polynomial features are
\[
    \varphi_\alpha(\mu)=\prod_{j=1}^3\sqrt{2\alpha_j+1}\,P_{\alpha_j}(\mu_j),
\]
where $P_k$ is the degree-$k$ Legendre polynomial. The factors normalize
the features under the uniform probability measure on the cube.
The rows, noisy labels, and constrained objective are defined in
\Cref{sec:experiments}. The degree-dependent coefficient bounds are
\[
    (r_0,r_1,r_2,r_3,r_4)=(0.15,0.02,10^{-3},8\cdot10^{-5},6\cdot10^{-6}).
\]
These chosen bounds favor lower-degree terms; they specify the
approximation class rather than a derived property of the PDE solution.
The sampling ellipsoid contains the box because
$x^\top Qx=\sum_{|\alpha|\le4}x_\alpha^2/(d r_{|\alpha|}^2)\le1$
whenever $|x_\alpha|\le r_{|\alpha|}$. The regression fits retain the
individual coefficient bounds.

\clearpage

\end{document}